%% file: template.tex
\documentclass{article}
\usepackage{footnote}
\makesavenoteenv{figure}
\usepackage{arxiv}
\usepackage[utf8]{inputenc}
\usepackage[T1]{fontenc}
\usepackage{CJKutf8}
\usepackage{mathptmx}
\usepackage{times}

\usepackage[dvipsnames]{xcolor}
\definecolor{kimiblue}{rgb}{0.09,0.5,0.99}
\usepackage[breaklinks,colorlinks,allcolors=kimiblue]{hyperref}
\usepackage{url}            % simple URL typesetting
\usepackage{booktabs}       % professional-quality tables
\usepackage{amsfonts}       % blackboard math symbols
\usepackage[normalem]{ulem}
\usepackage{xcolor}         % color support
\usepackage{nicefrac}       % compact symbols for 1/2, etc.
\usepackage{microtype}      % microtypography
\usepackage{textgreek}
\usepackage{graphicx}
\usepackage{listings}

\usepackage{doi}
\usepackage{algorithm}
\usepackage{algpseudocode}
\usepackage{multirow}
\usepackage{multicol}
\usepackage{xcolor}
\usepackage{subcaption}
\usepackage[normalem]{ulem}
\usepackage{amsmath,amsfonts,bm}
\usepackage{amssymb,amsthm}

\usepackage{enumitem}
\usepackage{subcaption}
\usepackage{adjustbox}
\usepackage{array}
\usepackage{makecell}
\usepackage{wrapfig}
\usepackage{ulem}
\usepackage[
    backend=biber,
    citestyle=numeric,
    bibstyle=numeric,
    mincitenames=1,
    maxcitenames=1,
    maxbibnames=3,
]{biblatex}
\newcommand{\citep}[1]{\parencite{#1}}
\setlist[itemize,1]{leftmargin=\dimexpr 18pt}
\setlist[enumerate,1]{leftmargin=\dimexpr 18pt}

\title{
\raisebox{-0.15\height}{\includegraphics[width=0.032\textwidth]{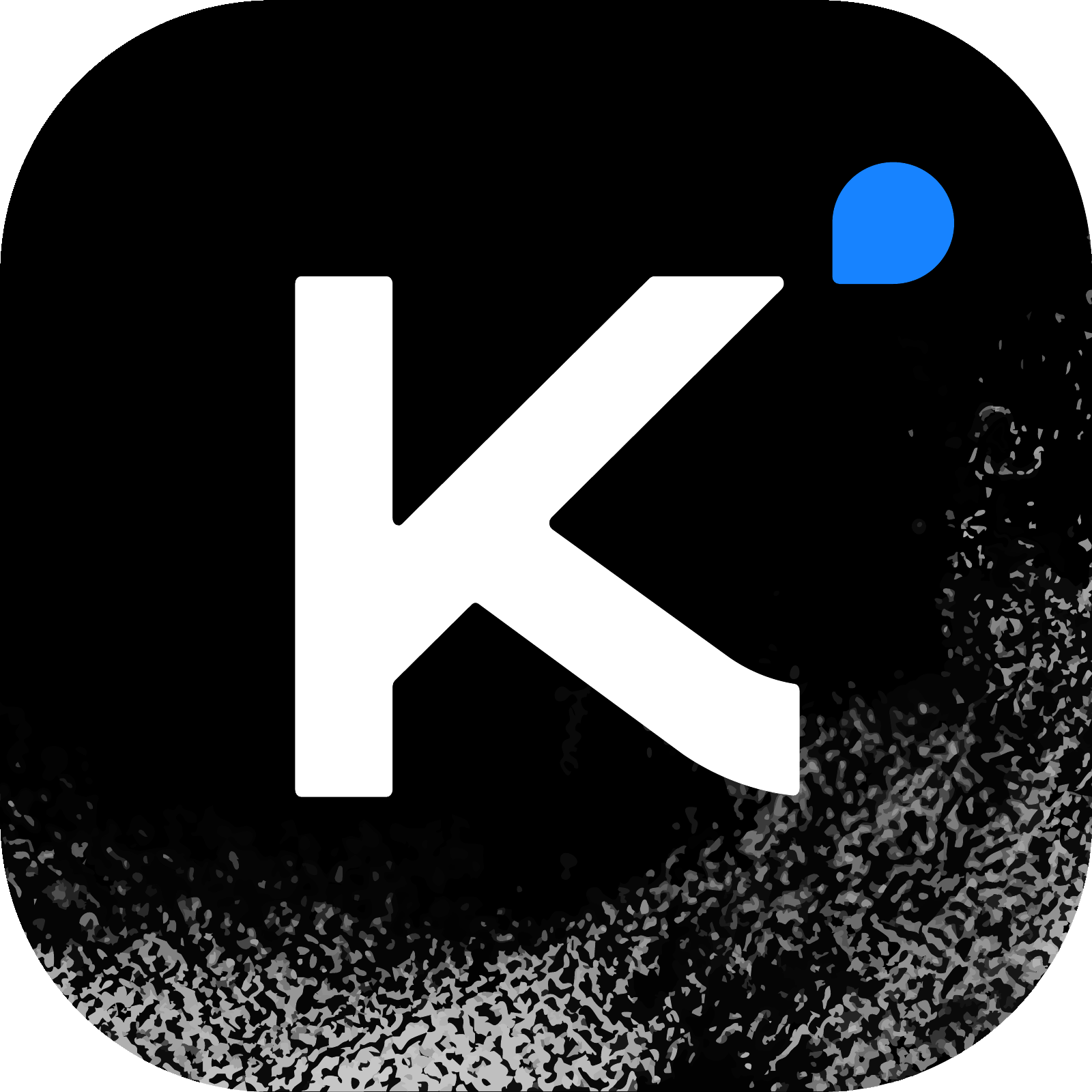}} %
Kimi K2: Open Agentic Intelligence}

\author{Kimi Team}

\date{}

\renewcommand{\headeright}{Technical Report}
\renewcommand{\undertitle}{Technical Report of Kimi K2}
\renewcommand{\shorttitle}{\raisebox{-0.12\height}{\includegraphics[width=0.02\textwidth]{figures/logo.pdf}}
Kimi K2}

\definecolor{darkblue}{rgb}{0.0, 0.0, 0.5}
\definecolor{darkgreen}{rgb}{0.0, 0.5, 0.0}
\definecolor{darkred}{rgb}{0.5, 0.0, 0.0}
\definecolor{darkpurple}{rgb}{0.5, 0.0, 0.5}

\newcommand{\chinese}[1]{\begin{CJK*}{UTF8}{gbsn}{#1}\end{CJK*}}

\begin{document}
\maketitle

\vspace{-12pt}
\begin{abstract}
We introduce Kimi K2, a Mixture-of-Experts (MoE) large language model with 32 billion activated parameters and 1 trillion total parameters.
We propose the MuonClip optimizer, which improves upon Muon with a novel QK-clip technique to address training instability while enjoying the advanced token efficiency of Muon. Based on MuonClip, K2 was pre-trained on 15.5 trillion tokens with zero loss spike.
During post-training, K2 undergoes a multi-stage post-training process, highlighted by a large-scale agentic data synthesis pipeline and a joint reinforcement learning (RL) stage, where the model improves its capabilities through interactions with real and synthetic environments.

Kimi K2 achieves state-of-the-art performance among open-source non-thinking models, with strengths in agentic capabilities. Notably, K2 obtains 66.1 on Tau2-Bench, 76.5 on ACEBench (En), 65.8 on SWE-Bench Verified, and 47.3 on SWE-Bench Multilingual --- surpassing most open and closed-sourced baselines in non-thinking settings.
It also exhibits strong capabilities in coding, mathematics, and reasoning tasks, with a score of 53.7 on LiveCodeBench v6, 49.5 on AIME 2025, 75.1 on GPQA-Diamond, and 27.1 on OJBench, all without extended thinking. These results position Kimi K2 as one of the most capable open-source large language models to date, particularly in software engineering and agentic tasks. We release our base and post-trained model checkpoints\footnote{\url{https://huggingface.co/moonshotai/Kimi-K2-Instruct}} to facilitate future research and applications of agentic intelligence.
\end{abstract}

\begin{figure}[htb]
    \centering
    \includegraphics[width=0.9\textwidth]{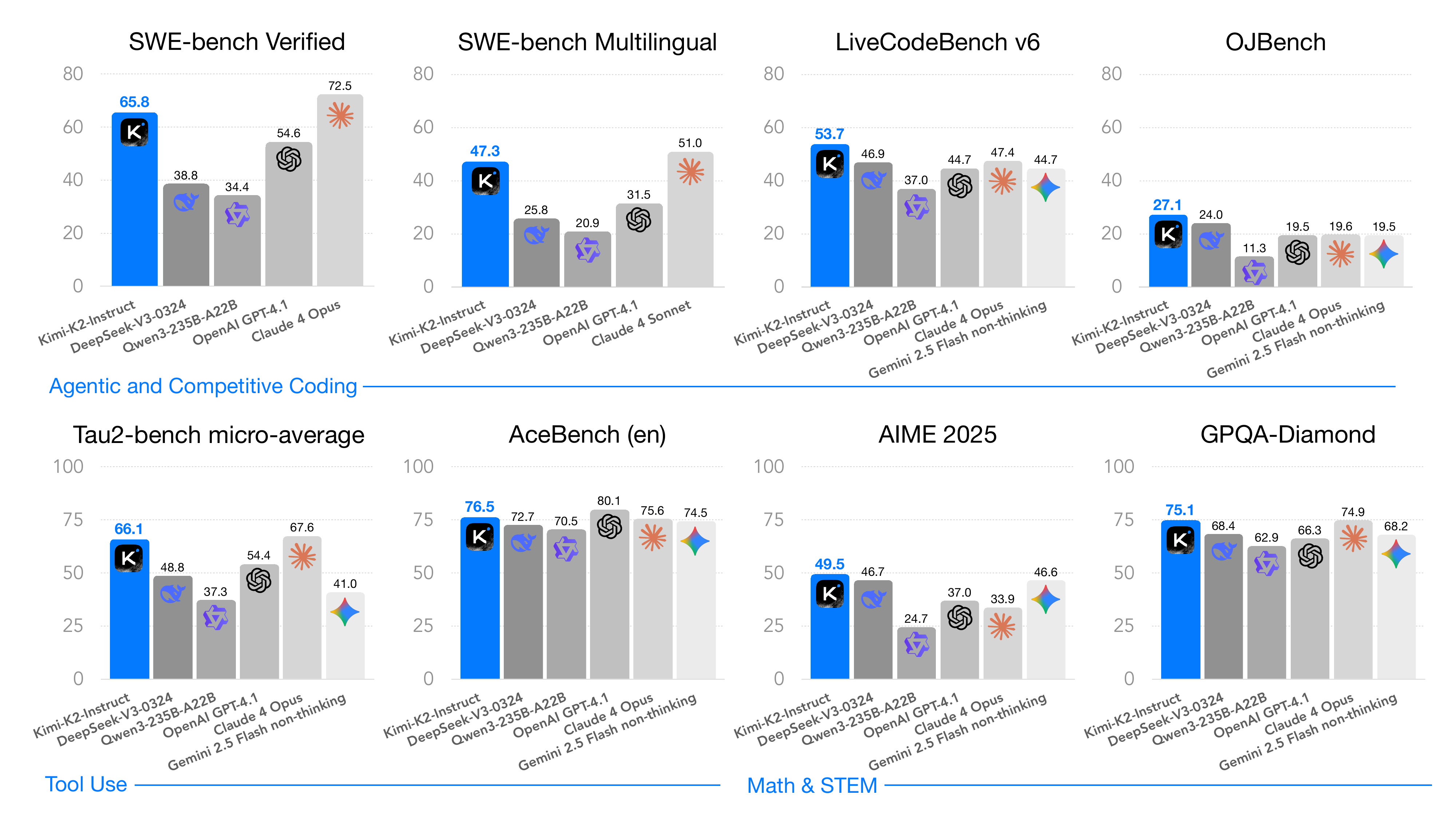}
    \caption{Kimi K2 main results.\protect\footnotemark} 
    \label{fig:kimi-k2-results}
\end{figure}
\footnotetext{All models evaluated above are non-thinking models. For SWE-bench Multilingual, we evaluated only Claude 4 Sonnet because the cost of Claude 4 Opus was prohibitive.}

\input{1-introduction}

\input{2-pretrain}

\input{3-post-train}
\input{4-results}

\input{6-conclusion}

\section{Acknowledgments}
We would like to acknowledge the valuable support provided by the OpenHands and Multi-SWE-bench teams in evaluating the SWE-bench Verified and Multi-SWE-bench experimental results.

\newpage
\printbibliography[title={References}]

\newpage
\appendix
\section*{Appendix}
\input{appendix}

\end{document}

%% file: 1-introduction.tex
\section{Introduction}

The development of Large Language Models (LLMs) is undergoing a profound paradigm shift towards \textit{Agentic Intelligence} -- the capabilities for models to autonomously perceive, plan, reason, and act within complex and dynamic environments. This transition marks a departure from static imitation learning towards models that actively learn through interactions, acquire new skills beyond their training distribution, and adapt behavior through experiences~\citep{silver2025welcome}. It is believed that this approach allows an AI agent to go beyond the limitation of static human-generated data, and acquire superhuman capabilities through its own exploration and exploitation. Agentic intelligence is thus rapidly emerging as a defining capability for the next generation of foundation models, with wide-ranging implications across tool use, software development, and real-world autonomy.

Achieving agentic intelligence introduces  challenges in both pre-training and post-training. Pre-training must endow models with broad general-purpose priors under constraints of limited high-quality data, elevating token efficiency—learning signal per token—as a critical scaling coefficient. Post-training must transform those priors into actionable behaviors, yet agentic capabilities such as multi-step reasoning, long-term planning, and tool use are rare in natural data and costly to scale. Scalable synthesis of structured, high-quality agentic trajectories, combined with general reinforcement learning (RL) techniques that incorporate preferences and self-critique, are essential to bridge this gap.

In this work, we introduce Kimi K2, a 1.04 trillion-parameter Mixture-of-Experts (MoE) LLM with 32 billion activated parameters, purposefully designed to address the core challenges and push the boundaries of agentic capability. Our contributions span both the pre-training and post-training frontiers:

\begin{itemize}
\item We present \textbf{MuonClip}, a novel optimizer that integrates the token-efficient Muon algorithm with a stability-enhancing mechanism called QK-Clip. Using MuonClip, we successfully pre-trained Kimi K2 on 15.5 trillion tokens without a single loss spike.

\item We introduce \textbf{a large-scale agentic data synthesis pipeline} that systematically generates tool-use demonstrations via simulated and real-world environments. This system constructs diverse tools, agents, tasks, and trajectories to create high-fidelity, verifiably correct agentic interactions at scale.

\item We design \textbf{a general reinforcement learning framework} that combines verifiable rewards (RLVR) with a self-critique rubric reward mechanism. The model learns not only from externally defined tasks but also from evaluating its own outputs, extending alignment from static into open-ended domains.
\end{itemize}

Kimi K2 demonstrates strong performance across a broad spectrum of agentic and frontier benchmarks. It achieves scores of 66.1 on Tau2-bench, 76.5 on ACEBench (en), 65.8 on SWE-bench Verified, and 47.3 on SWE-bench Multilingual, outperforming most open- and closed-weight baselines under non-thinking evaluation settings, closing the gap with Claude 4 Opus and Sonnet. In coding, mathematics, and broader STEM domains, Kimi K2 achieves 53.7 on LiveCodeBench v6, 27.1 on OJBench, 49.5 on AIME 2025, and 75.1 on GPQA-Diamond, further highlighting its capabilities in general tasks. On the LMSYS Arena leaderboard (July 17, 2025)\footnote{\url{https://lmarena.ai/leaderboard/text}}, Kimi K2 ranks as the top 1 open-source model and 5th overall based on over 3,000 user votes. 

To spur further progress in Agentic Intelligence, we are open-sourcing our base and post-trained checkpoints, enabling the community to explore, refine, and deploy agentic intelligence at scale.

%% file: 2-pretrain.tex
\section{Pre-training}

The base model of Kimi K2 is a trillion-parameter mixture-of-experts (MoE) transformer~\citep{transformer} model, pre-trained on 15.5 trillion high-quality tokens. Given the increasingly limited availability of high-quality human data, we posit that token efficiency is emerging as a critical coefficient in the scaling of large language models. To address this, we introduce a suite of pre-training techniques explicitly designed for maximizing token efficiency. Specifically, we employ the token-efficient Muon optimizer~\citep{jordan2024muon,liu2025muon} and mitigate its training instabilities through the introduction of QK-Clip. Additionally, we incorporate synthetic data generation to further squeeze the intelligence out of available high-quality tokens. The model architecture follows an ultra-sparse MoE with multi-head latent attention~(MLA) similar to DeepSeek-V3~\citep{deepseekai2024deepseekv3technicalreport} , derived from empirical scaling law analysis. The underlying infrastructure is built to optimize both training efficiency and research efficiency.

\subsection{MuonClip: Stable Training with Weight Clipping}

We train Kimi K2 using the token-efficient Muon optimizer~\citep{jordan2024muon}, incorporating weight decay and consistent update RMS scaling~\citep{liu2025muon}. Experiments in our previous work Moonlight~\citep{liu2025muon} show that, under the same compute budget and model size --- and therefore the same amount of training data --- Muon substantially outperforms AdamW~\citep{adam2015kingma, loshchilov2018decoupled}, making it an effective choice for improving token efficiency in large language model training.

\paragraph{Training instability when scaling Muon}
Despite its efficiency, scaling up Muon training reveals a challenge: training instability due to exploding attention logits, an issue that occurs more frequently with Muon but less with AdamW in our experiments. Existing mitigation strategies are insufficient. For instance, logit soft-cap~\citep{team2024gemma} directly clips the attention logits, but the dot products between queries and keys can still grow excessively before capping is applied. On the other hand, Query-Key Normalization (QK-Norm)~\citep{dehghani2023scaling,wortsman2309small} is not applicable to multi-head latent attention (MLA), because its Key matrices are not fully materialized during inference.

\paragraph{Taming Muon with QK-Clip}
To address this issue, we propose a novel weight-clipping mechanism \textit{QK-Clip} to explicitly constrain attention logits. QK-Clip works by rescaling the query and key projection weights post-update to bound the growth of attention logits.
 
Let the input representation of a transformer layer be $\mathbf X$. For each attention head $h$, its query, key, and value projections are computed as
\[
\mathbf{Q}^{h} = \mathbf X \mathbf W_q^{h}, \quad
\mathbf K^{h} = \mathbf X \mathbf W_k^{h}, \quad
\mathbf V^{h} = \mathbf X \mathbf W_v^{h}.
\]
where $\mathbf W_q, \mathbf W_k, \mathbf W_v$ are model parameters. The attention output is:

\[
\mathbf O^{h} = \operatorname{softmax}\left( \frac{1}{\sqrt{d}} \mathbf Q^{h} \mathbf K^{h\top} \right) \mathbf V^{h}.
\]

We define the max logit, a per-head scalar, as the maximum input to softmax in this batch $B$:
\[
S_{\max}^{h} = \frac{1}{\sqrt{d}} \max_{\mathbf X \in B} \max_{i,j} \mathbf Q_i^{h} \mathbf K_j^{h\top} 
\]
where $i,j$ are indices of different tokens in a training sample $\mathbf X$.

The core idea of QK-Clip is to rescale $\mathbf W_k, \mathbf W_q$ whenever $S_{\max}^{h}$ exceeds a target threshold $\tau$. Importantly, this operation does not alter the forward/backward computation in the current step --- we merely use the max logit as a guiding signal to determine the strength to control the weight growth.

A naïve implementation clips all heads at the same time:
\begin{equation*}
  \mathbf W_q^{h} \gets \gamma^{\alpha} \mathbf  W_q^{h} \qquad
  \mathbf W_k^{h} \gets  \gamma^{1-\alpha} \mathbf W_k^{h}
\end{equation*}
where $\gamma = \min(1, \tau / S_{\max})$ with $S_{\max} = \max_h S_{\max}^{h}$, and $\alpha$ is a balancing parameter typically set to $0.5$, applying equal scaling to queries and keys. 

However, we observe that in practice, only a small subset of heads exhibit exploding logits. In order to minimize our intervention on model training, we determine a per-head scaling factor $\gamma_h = \min(1, \tau / S_{\max}^h) $, and opt to apply per-head QK-Clip. Such clipping is straightforward for regular multi-head attention (MHA). For MLA, we apply clipping only on unshared attention head components:

\begin{itemize}
\item $\textbf{q}^{C}$ and $\textbf{k}^{C}$ (head-specific components): each scaled by $\sqrt{\gamma_h}$ 
\item $\textbf{q}^{R}$ (head-specific rotary): scaled by $\gamma_h$,
\item $\textbf{k}^{R}$ (shared rotary): left untouched to avoid effect across heads.
\end{itemize}

\begin{algorithm}[t]
\caption{MuonClip Optimizer}
\label{alg:muonclip}
\begin{algorithmic}[1]
\For{each training step $t$}
   \State \textcolor{gray}{// 1. Muon optimizer step}
   \For{each weight $\mathbf W\in\mathbb{R}^{n\times m}$}
        \State $\mathbf M_t = \mu \mathbf M_{t-1} + \mathbf G_t$ \Comment{$\mathbf M_0 = \mathbf 0$, $\mathbf G_t$ is the grad of $\mathbf W_t$, $\mu$ is momentum}
        \State $\mathbf O_t = \operatorname{Newton-Schulz}(\mathbf M_t)\cdot \sqrt{\max(n,m)}\cdot 0.2$ \Comment{Match Adam RMS}
        \State $\mathbf W_t = \mathbf W_{t-1} - \eta \bigl(\mathbf O_t + \lambda \mathbf W_{t-1}\bigr)$ \Comment{learning rate $\eta$, weight decay $\lambda$}
   \EndFor
   \State \textcolor{gray}{// 2. QK-Clip}
   \For{each attention head $h$ in every attention layer of the model}
       \State Obtain $S_{\max}^{h}$ already computed during forward
       \If{$S_{\max}^{h} > \tau$}
           \State $\gamma \gets \tau / S_{\max}^{h}$
           \State $\mathbf W_{qc}^{h} \gets \mathbf W_{qc}^{h} \cdot \sqrt{\gamma}$
           \State $\mathbf W_{kc}^{h} \gets \mathbf W_{kc}^{h} \cdot \sqrt{\gamma}$
           \State $\mathbf W_{qr}^{h} \gets \mathbf W_{qr}^{h} \cdot \gamma$
       \EndIf
   \EndFor
\EndFor
\end{algorithmic}
\end{algorithm}

\begin{figure}[htbp]
  \centering
  \includegraphics[width=.48\linewidth]{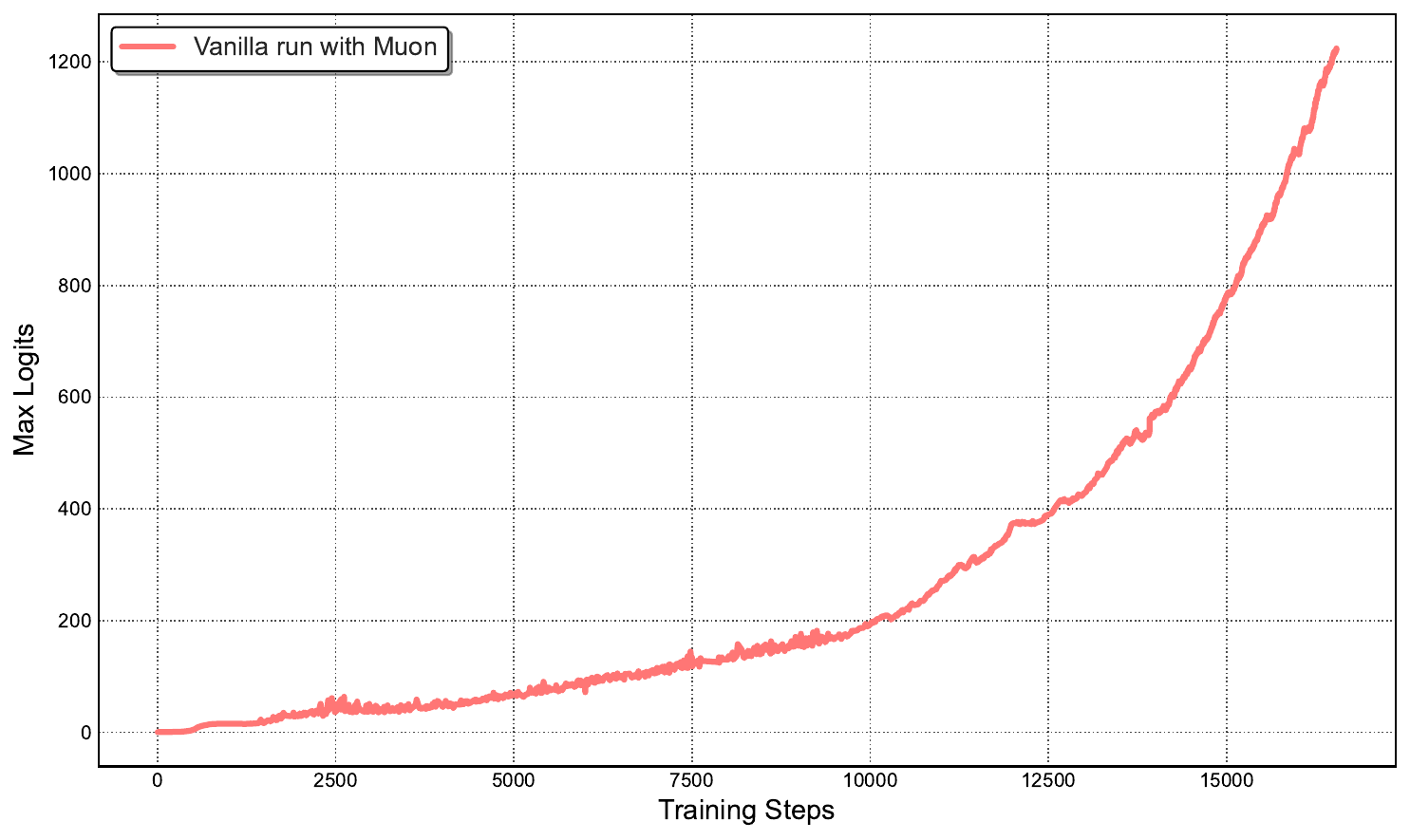}
  \hfill
  \includegraphics[width=.48\linewidth]{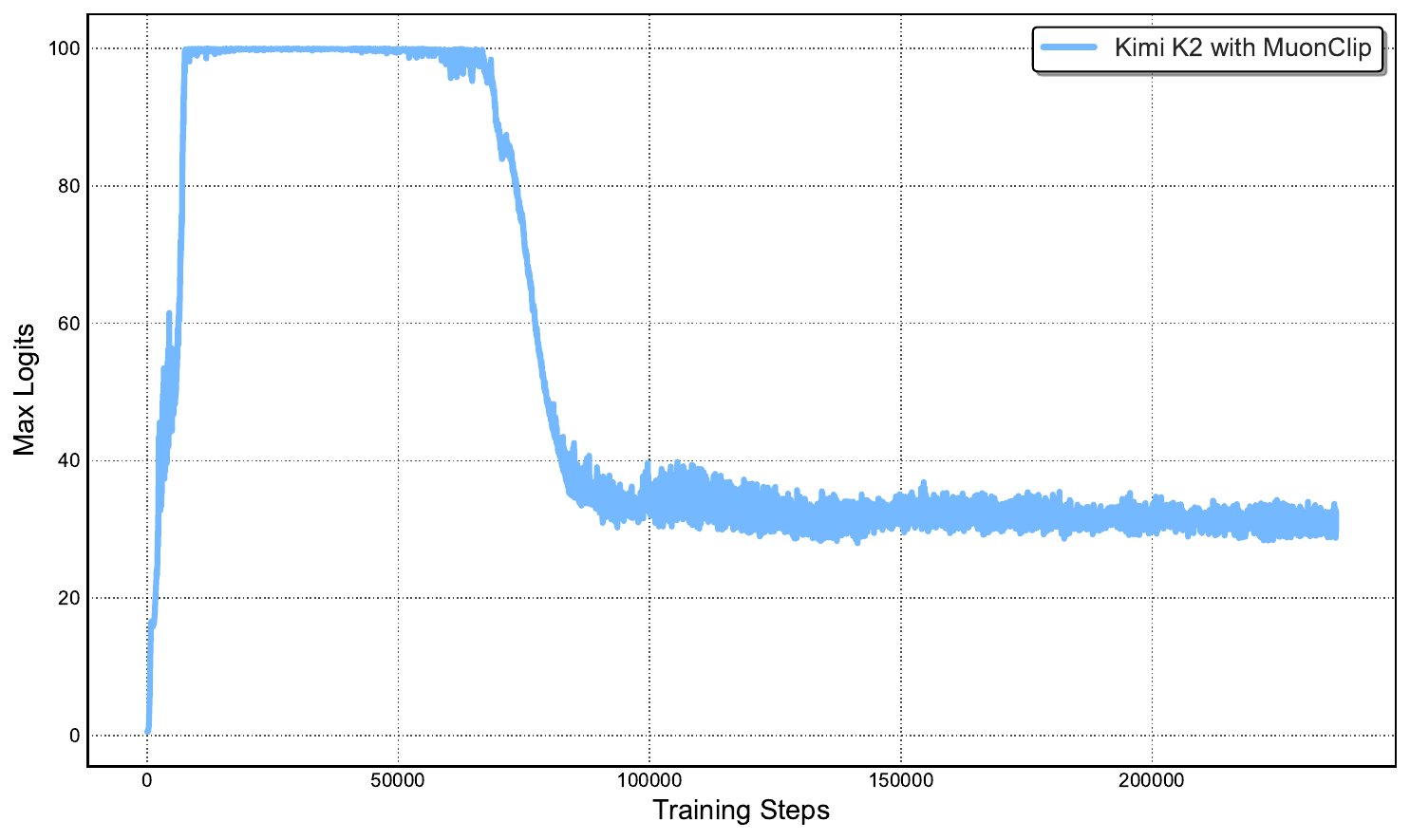}

  \caption{Left: During a mid-scale training run, attention logits rapidly exceed 1000, which could lead to potential numerical instabilities and even training divergence.  Right: Maximum logits for Kimi K2 with MuonClip and $\tau$ = 100 over the entire training run. 
  The max logits rapidly increase to the capped value of 100, and only decay to a stable range after approximately 30\% of the training steps, demonstrating the effective regulation effect of QK-Clip.}

  \label{fig:mini_muonclip}
\end{figure}

\paragraph{MuonClip: The New Optimizer}

We integrate Muon with weight decay, consistent RMS matching, and QK-Clip into a single optimizer, which we refer to as \textbf{MuonClip} (see Algorithm~\ref{alg:muonclip}).

We demonstrate the effectiveness of MuonClip from several scaling experiments. First, we train a mid-scale 9B activated and 53B total parameters Mixture-of-Experts (MoE) model using the vanilla Muon. As shown in Figure~\ref{fig:mini_muonclip} (Left), we observe that the maximum attention logits quickly exceed a magnitude of 1000, showing that attention logits explosion is already evident in Muon training to this scale. Max logits at this level usually result in instability during training, including significant loss spikes and occasional divergence.

Next, we demonstrate that QK-Clip does not degrade model performance and confirm that the MuonClip optimizer preserves the optimization characteristics of Muon without adversely affecting the loss trajectory. A detailed discussion of the experiment designs and findings is provided in the Appendix~\ref{sec:qkclip_harmless}.

\begin{figure}[htb]
    \centering
    \includegraphics[width=0.7\textwidth]{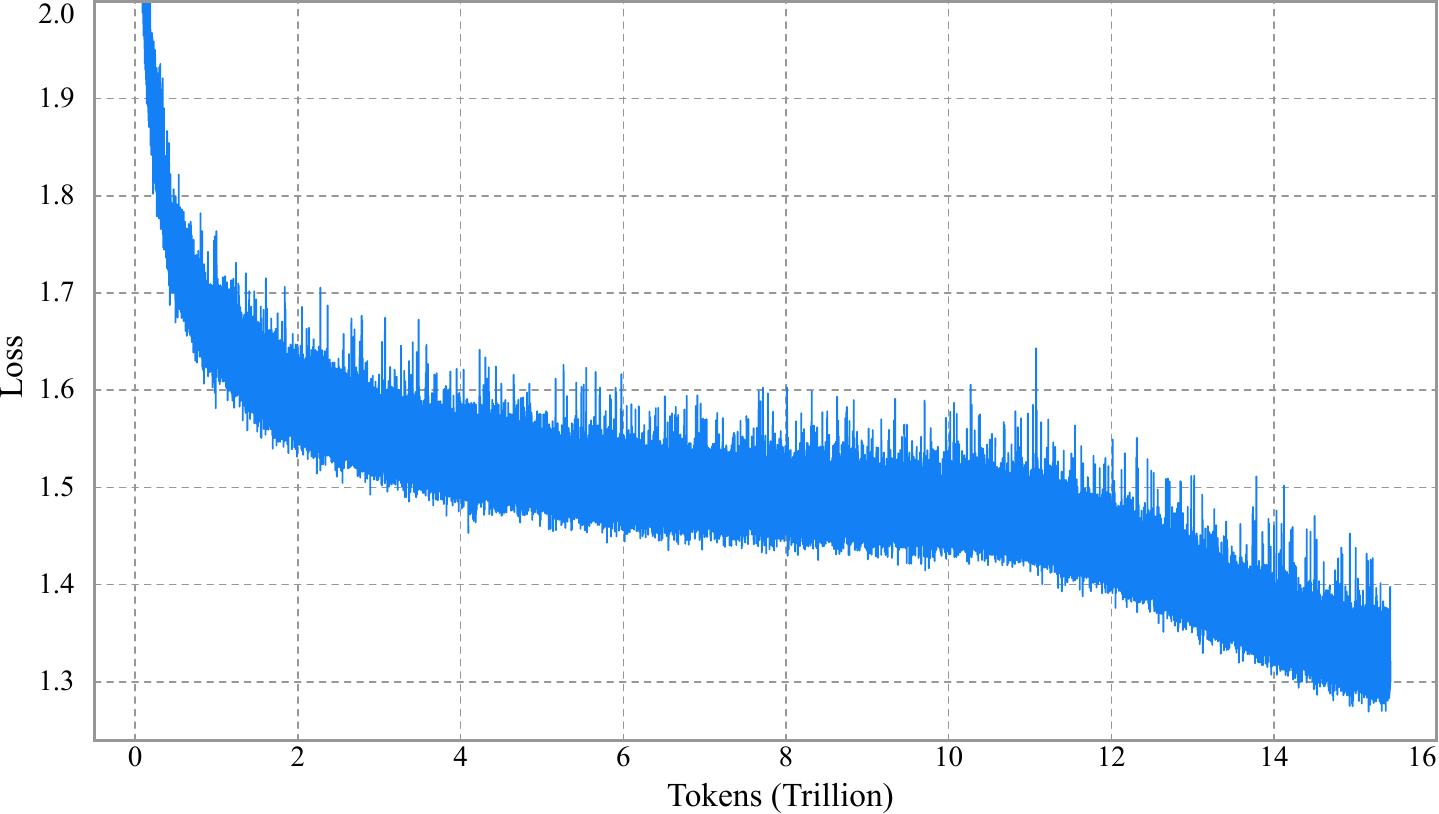}
    \caption{Per-step training loss curve of Kimi K2, \textbf{without smoothing or sub-sampling}. It shows no spikes throughout the entire training process. Note that we omit the very beginning of training for clarity.} 
    \label{fig:kimi-loss-curve}
\end{figure}

Finally, we train Kimi K2, a large-scale MoE model, using MuonClip with $\tau = 100$ and monitor the maximum attention logits throughout the training run (Figure~\ref{fig:mini_muonclip} (Right)). Initially, the logits are capped at 100 due to QK-Clip. Over the course of training, the maximum logits gradually decay to a typical operating range without requiring any adjustment to $\tau$. Importantly, the training loss remains smooth and stable, with no observable spikes, as shown in Figure~\ref{fig:kimi-loss-curve}, validating that MuonClip provides robust and scalable control over attention dynamics in large-scale language model training.

\subsection{Pre-training Data: Improving Token Utility with Rephrasing}

Token efficiency in pre-training refers to how much performance improvement is achieved for each token consumed during training. Increasing token utility—the effective learning signal each token contributes—enhances the per-token impact on model updates, thereby directly improving token efficiency. 
This is particularly important when the supply of high-quality tokens is limited and must be maximally leveraged. A naive approach to increasing token utility is through repeated exposure to the same tokens, which can lead to overfitting and reduced generalization.

A key advancement in the pre-training data of Kimi K2 over Kimi K1.5 is the introduction of a synthetic data generation strategy to increase token utility. Specifically, a carefully designed rephrasing pipeline is employed to amplify the volume of high-quality tokens without inducing significant overfitting. In this report, we describe two domain-specialized rephrasing techniques—targeted respectively at the Knowledge and Mathematics domains—that enable this controlled data augmentation.

\begin{figure}[t]
\centering
\begin{minipage}[t]{0.8\textwidth}
  \centering
  \includegraphics[width=\linewidth]{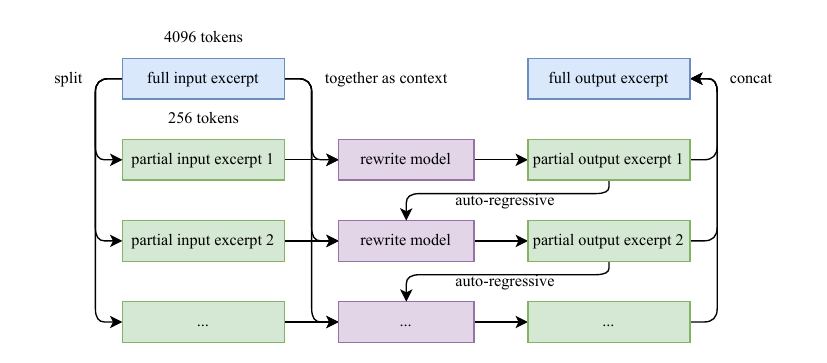}
  \caption{Auto-regressive chunk-wise rephrasing pipeline for long input excerpts. The input is split into smaller chunks with preserved context, rewritten sequentially, and then concatenated into a full rewritten passage. }
  \label{fig:rephrase_pipeline}
\end{minipage}\hfill
\end{figure}

\paragraph{Knowledge Data Rephrasing}
Pre-training on natural, knowledge-intensive text presents a trade-off: a single epoch is insufficient for comprehensive knowledge absorption, while multi-epoch repetition yields diminishing returns and increases the risk of overfitting. To improve the token utility of high-quality knowledge tokens, we propose a synthetic rephrasing framework composed of the following key components:

\begin{itemize}
    \item \textbf{Style- and perspective-diverse prompting:} Inspired by WRAP~\citep{maini2024rephrasingwebrecipecompute}, we apply a range of carefully engineered prompts to enhance linguistic diversity while maintaining factual integrity. These prompts guide a large language model to generate faithful rephrasings of the original texts in varied styles and from different perspectives.
    
    \item \textbf{Chunk-wise autoregressive generation:} To preserve global coherence and avoid information loss in long documents, we adopt a chunk-based autoregressive rewriting strategy. Texts are divided into segments, rephrased individually, and then stitched back together to form complete passages. This method mitigates implicit output length limitations that typically exist with LLMs. An overview of this pipeline is presented in Figure~\ref{fig:rephrase_pipeline}.

    \item \textbf{Fidelity verification:} To ensure consistency between original and rewritten content, we perform fidelity checks that compare the semantic alignment of each rephrased passage with its source. This serves as an initial quality control step prior to training.
\end{itemize}

We compare data rephrasing with multi-epoch repetition by testing their corresponding accuracy on SimpleQA.
We experiment with an early checkpoint of K2 and evaluate three training strategies: (1) repeating the original dataset for 10 epochs, (2) rephrasing the data once and repeating it for 10 epochs, and (3) rephrasing the data 10 times with a single training pass. 
As shown in Table~\ref{tab:rewrites_epochs_f1}, the accuracy consistently improves across these strategies, demonstrating the efficacy of our rephrasing-based augmentation. We extended this method to other large-scale knowledge corpora and observed similarly encouraging results, and each corpora is rephrased at most twice.

\begin{table}[htbp]
\centering
\caption{SimpleQA Accuracy under three rephrasing-epoch configurations}
\label{tab:rewrites_epochs_f1}
\begin{tabular}{ccc}
\toprule
\# Rephrasings & \# Epochs & SimpleQA Accuracy \\
\midrule
0 (raw wiki-text) & 10 & 23.76 \\
1            & 10 & 27.39 \\
10           & 1  & 28.94 \\
\bottomrule
\end{tabular}
\end{table}

\paragraph{Mathematics Data Rephrasing} To enhance mathematical reasoning capabilities, we rewrite high-quality mathematical documents into a ``learning-note'' style, following the methodology introduced in SwallowMath~\citep{fujii2025rewritingpretrainingdataboosts}. In addition, we increased data diversity by translating high-quality mathematical materials from other languages into English.

Although initial experiments with rephrased subsets of our datasets show promising results, the use of synthetic data as a strategy for continued scaling remains an active area of investigation. Key challenges include generalizing the approach to diverse source domains without compromising factual accuracy, minimizing hallucinations and unintended toxicity, and ensuring scalability to large-scale datasets.

\paragraph{Pre-training Data Overall} The Kimi K2 pre-training corpus comprises 15.5 trillion tokens of curated, high-quality data spanning four primary domains: Web Text, Code, Mathematics, and Knowledge. Most data processing pipelines follow the methodologies outlined in Kimi K1.5~\citep{team2025kimi}. For each domain, we performed rigorous correctness and quality validation and designed targeted data experiments to ensure the curated dataset achieved both high diversity and effectiveness.

\subsection{Model Architecture}
Kimi K2 is a 1.04 trillion-parameter Mixture-of-Experts (MoE) transformer model with 32 billion activated parameters. The architecture follows a similar design to DeepSeek-V3~\citep{deepseekai2024deepseekv3technicalreport} , employing Multi-head Latent Attention (MLA)~\citep{liu2024deepseek} as the attention mechanism, with a model hidden dimension of 7168 and an MoE expert hidden dimension of 2048. Our scaling law analysis reveals that continued increases in sparsity yield substantial performance improvements, which motivated us to increase the number of experts to 384, compared to 256 in DeepSeek-V3. To reduce computational overhead during inference, we cut the number of attention heads to 64, as opposed to 128 in DeepSeek-V3. Table~\ref{tab:k2_arch_comparison} presents a detailed comparison of architectural parameters between Kimi K2 and DeepSeek-V3.

\begin{table}[htbp]
\centering
\caption{Architectural comparison between Kimi K2 and DeepSeek-V3}
\label{tab:k2_arch_comparison}
\begin{tabular}{lccc}
\toprule
  & \textbf{DeepSeek-V3} & \textbf{Kimi K2} & \textbf{$\Delta$} \\
\midrule
\#Layers & 61 & 61 & = \\
Total Parameters & 671B & 1.04T & $\uparrow$ 54\% \\
Activated Parameters & 37B & 32.6B & $\downarrow$ 13\% \\
Experts (total) & 256 & 384 & $\uparrow$ 50\% \\
Experts Active per Token & 8 & 8 & = \\
Shared Experts & 1 & 1 & = \\
Attention Heads & 128 & 64 & $\downarrow$ 50\% \\
Number of Dense Layers & 3 & 1 & $\downarrow$ 67\% \\
Expert Grouping & Yes & No  & - \\
\bottomrule
\end{tabular}
\end{table}

\begin{figure}[htbp]
\centering
\begin{minipage}[t]{0.48\textwidth}
  \centering

  \includegraphics[width=\linewidth]{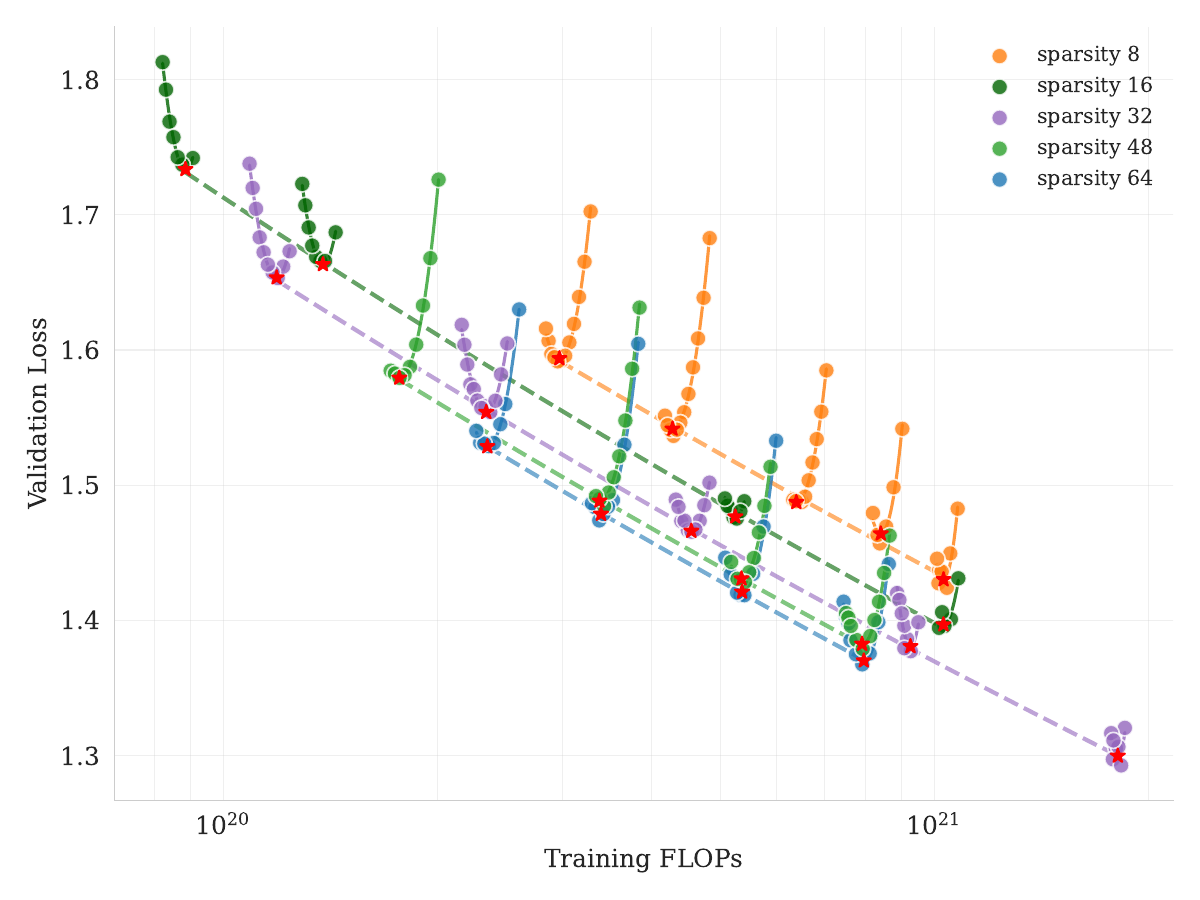}
  \caption{Sparsity Scaling Law. Increasing sparsity leads to improved model performance. We fixed the number of activated experts to 8 and the number of shared experts to 1, and varied the total number of experts, resulting in models with different sparsity levels.}
  \label{fig:scaling-law}

\end{minipage}\hfill
\begin{minipage}[t]{0.48\textwidth}
  \centering

  \includegraphics[width=\linewidth]{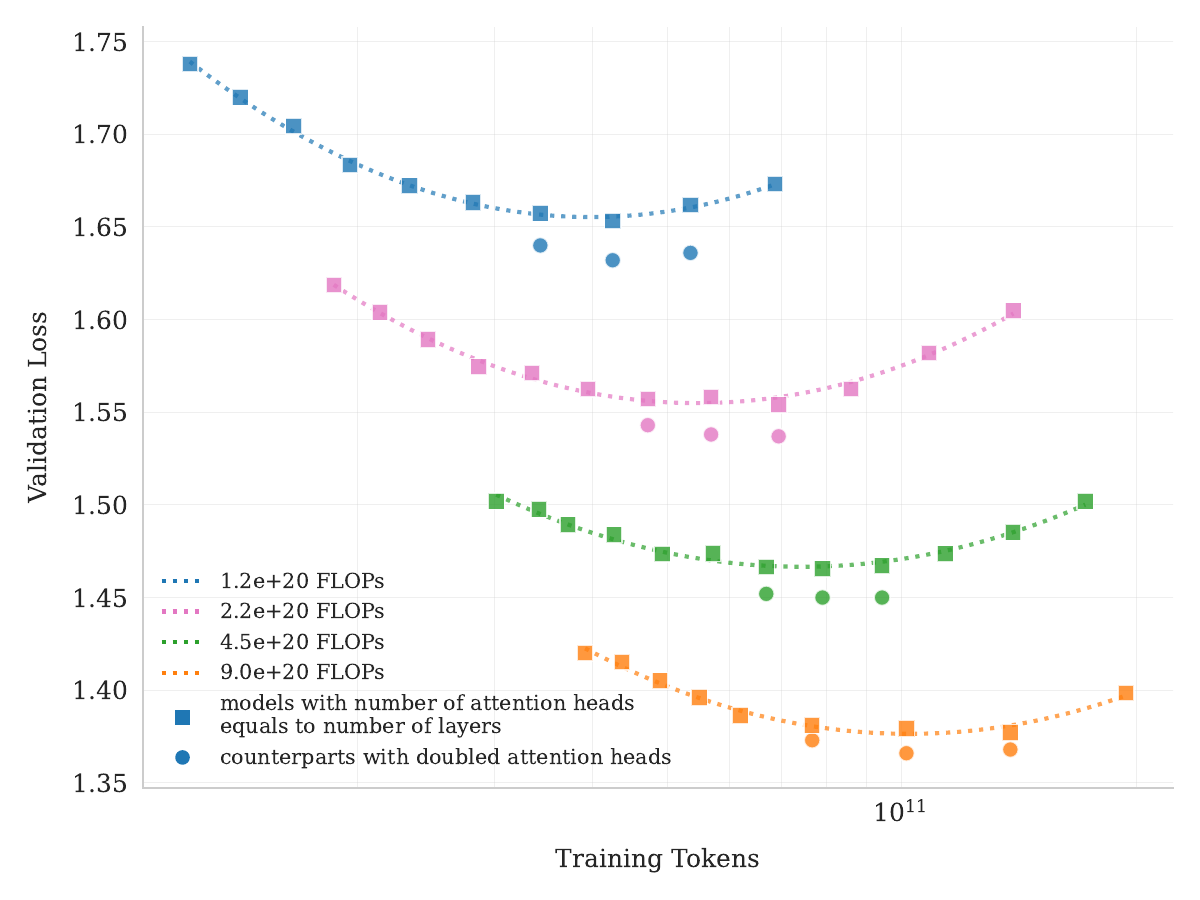}
  \caption{Scaling curves for models with number of attention heads equals to number of layers and their counterparts with doubled attention heads. Doubling the number of attention heads leads to a reduction in validation loss of approximately $0.5\%$ to $1.2\%$.}
  \label{fig:doubleQheads}
 
\end{minipage}
\end{figure}

\paragraph{Sparsity Scaling Law}
We develop a sparsity scaling law tailored for the Mixture-of-Experts (MoE) model family using Muon. Sparsity is defined as the ratio of the total number of experts to the number of activated experts. Through carefully controlled small-scale experiments, we observe that --- under a fixed number of activated parameters (i.e., constant FLOPs) --- increasing the total number of experts (i.e., increasing sparsity) consistently lowers both the training and validation loss, thereby enhancing overall model performance (Figure~\ref{fig:scaling-law}). Concretely, under the compute-optimal sparsity scaling law, achieving the same validation loss of 1.5, sparsity 48 reduces FLOPs by 1.69×, 1.39×, and 1.15× compared to sparsity levels 8, 16, and 32, respectively.
Though increasing sparsity leads to better performance, this gain comes with increased infrastructure complexity. To balance model performance with cost, we adopt a sparsity of 48 for Kimi K2, activating 8 out of 384 experts per forward pass.

\paragraph{Number of Attention Heads}

DeepSeek-V3~\citep{deepseekai2024deepseekv3technicalreport}  sets the number of attention heads to roughly twice the number of model layers to better utilize memory bandwidth and enhance computational efficiency. 
However, as the context length increases, doubling the number of attention heads leads to significant inference overhead, reducing efficiency at longer sequence lengths. This becomes a major limitation in agentic applications, where efficient long context processing is essential.
For example, with a sequence length of 128k, increasing the number of attention heads from 64 to 128, while keeping the total expert count fixed at 384, leads to an 83\% increase in inference FLOPs.
To evaluate the impact of this design, we conduct controlled experiments comparing configurations where the number of attention heads equals the number of layers against those with double number of heads, under varying training FLOPs. Under iso-token training conditions, we observe that doubling the attention heads yields only modest improvements in validation loss (ranging from 0.5\% to 1.2\%) across different compute budgets (Figure~\ref{fig:doubleQheads}).
Given that sparsity 48 already offers strong performance, the marginal gains from doubling attention heads do not justify the inference cost. Therefore we choose to 64 attention heads.

\subsection{Training Infrastructure}
\subsubsection{Compute Cluster}

Kimi K2 was trained on a cluster equipped with NVIDIA H800 GPUs. Each node in
the H800 cluster contains 2 TB RAM and 8 GPUs connected by NVLink and NVSwitch within nodes. Across
different nodes, $\text{8}\!\times\!\text{400}~\text{Gbps}$ RoCE interconnects are utilized to facilitate communications.

\subsubsection{Parallelism for Model Scaling}

Training of large language models often progresses under dynamic resource availability. Instead of optimizing one parallelism strategy that's only applicable under specific amount of resources, we pursue a flexible strategy that allows Kimi K2 to be trained on any number of nodes that is a multiple of 32.
Our strategy leverages a combination of 16-way Pipeline Parallelism (PP) with virtual stages~\citep{huang2019gpipe, narayanan2021efficient, lamy2023breadth, qi2023zero, Liu_2023, harlap2018pipedream}, 16-way Expert Parallelism (EP)~\citep{lepikhin2020gshard}, and ZeRO-1 Data Parallelism~\citep{rajbhandari2020zero}.

Under this setting, storing the model parameters in BF16 and their gradient accumulation buffer in FP32 requires approximately 6 TB of GPU memory, distributed over a model-parallel group of 256 GPUs. 
Placement of optimizer states depends on the training configurations. When the total number of training nodes is large, the optimizer states are distributed, reducing its per-device memory footprint to a negligible level. When the total number of training nodes is small (e.g., 32), we can offload some optimizer states to CPU. 

This approach allows us to reuse an identical parallelism configuration for both small- and large-scale experiments, while letting each GPU hold approximately 30 GB of GPU memory for all states. The rest of the GPU memory are used for activations, as described in Sec.~\ref{sec:act_reduction}.
Such a consistent design is important for research efficiency, as it simplifies the system and substantially accelerates experimental iteration.

\paragraph{EP communication overlap with interleaved 1F1B}

By increasing the number of warm-up micro-batches, we can overlap EP all-to-all communication with computation under the standard interleaved 1F1B schedule \citep{harlap2018pipedream, narayanan2021efficient}. 
In comparison, DualPipe~\citep{deepseekai2024deepseekv3technicalreport} doubles the memory required for parameters and gradients, necessitating an increase in parallelism to compensate.
Increasing PP introduces more bubbles, while increasing EP, as discussed below, incurs higher overhead.
The additional costs are prohibitively high for training a large model with over 1 trillion parameters and thus we opted not to use DualPipe. 

However, interleaved 1F1B splits the model into more stages, introducing non-trivial PP communication overhead. To mitigate this cost, we decouple the weight-gradient computation from each micro-batch's backward pass and execute it in parallel with the corresponding PP communication.
Consequently, all PP communications can be effectively overlapped except for the warm-up phase.

\paragraph{Smaller EP size}

To ensure full computation-communication overlap during the 1F1B stage, the reduced attention computation time in K2~(which has 64 attention heads compared to 128 heads in DeepSeek-V3) necessitates minimizing the time of EP operations. 
This is achieved by adopting the smallest feasible EP parallelization strategy, specifically EP = 16.
Utilizing a smaller EP group also relaxes expert-balance constraints, allowing for near-optimal speed to be achieved without further tuning.

\subsubsection{Activation Reduction}
\label{sec:act_reduction}

After reserving space for parameters, gradient buffers, and optimizer states, the remaining GPU memory on each device is insufficient to hold the full MoE activations. To ensure the activation memory fits within the constraints, especially for the initial pipeline stages that accumulate the largest activations during the 1F1B warm-up phase, the following techniques are employed.

\paragraph{Selective recomputation}

Recomputation is applied to inexpensive, high-footprint stages, including \texttt{LayerNorm}, \texttt{SwiGLU}, and \texttt{MLA} up-projections~\citep{deepseekai2024deepseekv3technicalreport}.
Additionally, \texttt{MoE} down-projections are recomputed during training to further reduce activation memory. While optional, this recomputation maintains adequate GPU memory, preventing crashes caused by expert imbalance in early training stages.

\paragraph{FP8 storage for insensitive activations}

Inputs of \texttt{MoE} up-projections and \texttt{SwiGLU} are compressed to FP8-E4M3 in 1$\times$ 128 tiles with FP32 scales. Small-scale experiments show no measurable loss increase. Due to potential risks of performance degradation that we observed during preliminary study, we do not apply FP8 in computation.

\begin{figure}
    \centering
    \includegraphics[width=1\linewidth]{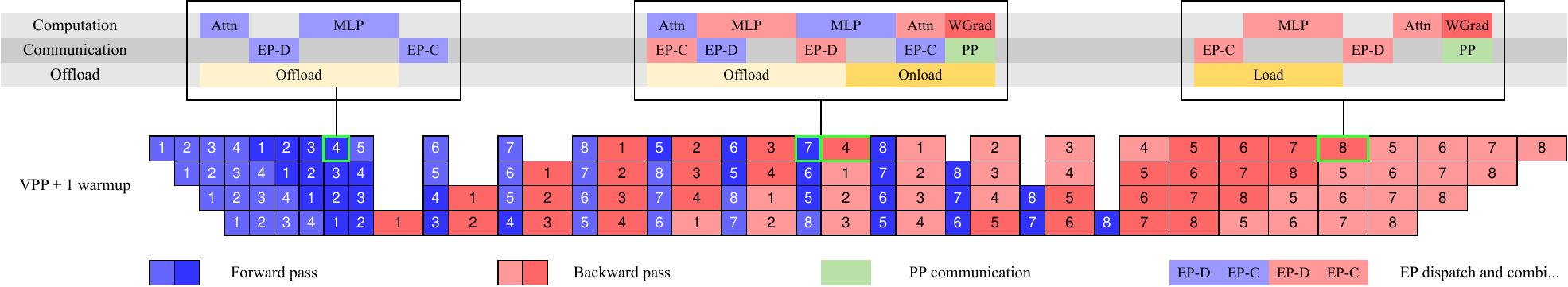}
    \caption{Computation, communication and offloading overlapped in different PP phases.}
    \label{fig:3x_stream}
\end{figure}
\paragraph{Activation CPU offload}

All remaining activations are offloaded to CPU RAM. A copy engine is responsible for streaming the offload and onload, overlapping with both computation and communication kernels. During the 1F1B phase, we offload the forward activations of the previous micro-batch while prefetching the backward activations of the next. The warm-up and cool-down phases are handled similarly and the overall pattern is shown in Figure~\ref{fig:3x_stream}. Although offloading may slightly affect EP traffic due to PCIe traffic congestion, our tests show that EP communication remains fully overlapped.

\subsection{Training recipe}

We pre-trained the model with a 4,096-token context window using the MuonClip optimizer (Algorithm~\ref{alg:muonclip}) and the WSD learning rate schedule~\citep{hu2024minicpm}, processing a total of 15.5T tokens. 
The first 10T tokens were trained with a constant learning rate of 2e-4 after a 500-step warm-up,
followed by 5.5T tokens with a cosine decay from 2e-4 to 2e-5. 
Weight decay was set to 0.1 throughout, and the global batch size was held at 67M tokens. The overall training curve is shown in Figure~\ref{fig:kimi-loss-curve}.

Towards the end of pre-training, we conducted an annealing phase followed by a long-context activation stage. The batch size was kept constant at 67M tokens, while the learning rate was decayed from 2e-5 to 7e-6. In this phase, the model was trained on 400 billion tokens with a 4k sequence length, followed by an additional 60 billion tokens with a 32k sequence length. To extend the context window to 128k, we employed the YaRN method~\citep{peng2023yarn}.

%% file: 3-post-train.tex
\section{Post-Training}

\subsection{Supervised Fine-Tuning}

We employ the Muon optimizer~\citep{jordan2024muon} in our post-training and recommend its use for fine-tuning with K2. This follows from the conclusion of our previous work~\cite{liu2025muon} that a Muon-pre-trained checkpoint produces the best performance with Muon fine-tuning.

We construct a large-scale instruction-tuning dataset spanning diverse domains, guided by two core principles: maximizing prompt diversity and ensuring high response quality. To this end, we develop a suite of data generation pipelines tailored to different task domains, each utilizing a combination of human annotation, prompt engineering, and verification processes.
We adopt K1.5~\citep{team2025kimi} and other in-house domain-specialized expert models to generate candidate responses for various tasks, followed by LLMs or human-based judges to perform automated quality evaluation and filtering.
For agentic data, we create a data synthesis pipeline to teach models tool-use capabilities through multi-step, interactive reasoning.

\subsubsection{Large-Scale Agentic Data Synthesis for Tool Use Learning}
\label{sec:agentic-data-synthesis}

A critical capability of modern LLM agents is their ability to autonomously use unfamiliar tools, interact with external environments, and iteratively refine their actions through reasoning, execution, and error correction. Agentic tool use capability is essential for solving complex, multi-step tasks that require dynamic interaction with real-world systems. Recent benchmarks such as ACEBench~\citep{chen2025acebench} and $\tau$-bench~\citep{yao2024tau} have highlighted the importance of comprehensive tool-use evaluation, while frameworks like ToolLLM~\citep{qin2023toolllm} and ACEBench~\citep{chen2025acebench} have demonstrated the potential of teaching models to use thousands of tools effectively.

However, training such capabilities at scale presents a significant challenge: while real-world environments provide rich and authentic interaction signals, they are often difficult to construct at scale due to cost, complexity, privacy and accessibility constraints. Recent work on synthetic data generation (AgentInstruct~\citep{mitra2024agentinstruct}; Self-Instruct~\citep{wang2022self}; StableToolBench~\citep{guo2025stabletoolbenchstablelargescalebenchmarking}; ZeroSearch~\citep{sun2025zerosearchincentivizesearchcapability}) has shown promising results in creating large-scale data without relying on real-world interactions. Building on these advances and inspired by ACEBench~\citep{chen2025acebench}'s comprehensive data synthesis framework, we developed a pipeline that simulates real-world tool-use scenarios at scale, enabling the generation of tens of thousands of diverse and high-quality training examples.

\begin{figure}[t]
    \centering
    \begin{subfigure}[b]{0.4\textwidth}
        \centering
        \includegraphics[width=\textwidth]{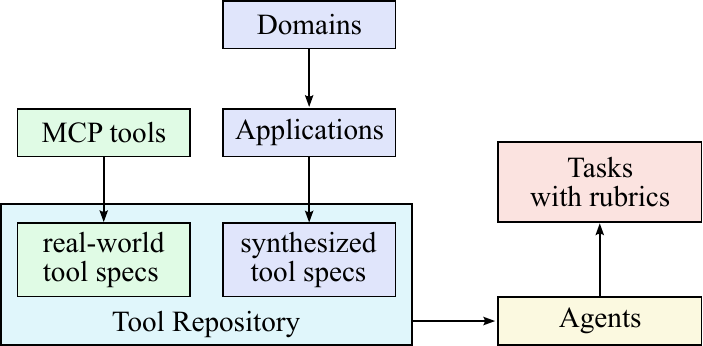}
        \caption{Synthesizing tool specs, agents and tasks}
        \label{fig:tool_repo_synth}
    \end{subfigure}
    \qquad
    \begin{subfigure}[b]{0.4\textwidth}
        \centering
        \includegraphics[width=\textwidth]{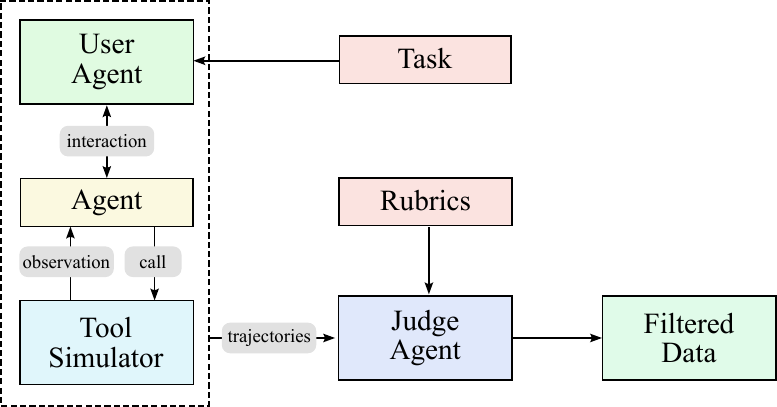}
        \caption{Generating agent trajectories}
        \label{fig:tool_traj_synth}
    \end{subfigure}
    \caption{Data synthesis pipeline for tool use. (a) Tool specs are from both real-world tools and LLMs; agents and tasks are the generated from the tool repo. (b) Multi-agent pipeline to generate and filter trajectories with tool calling.}
    \label{fig:tool_synth}
\end{figure}

\begin{figure}[t]
    \centering
    \begin{subfigure}[b]{0.48\textwidth}
        \centering
        \includegraphics[width=\textwidth]{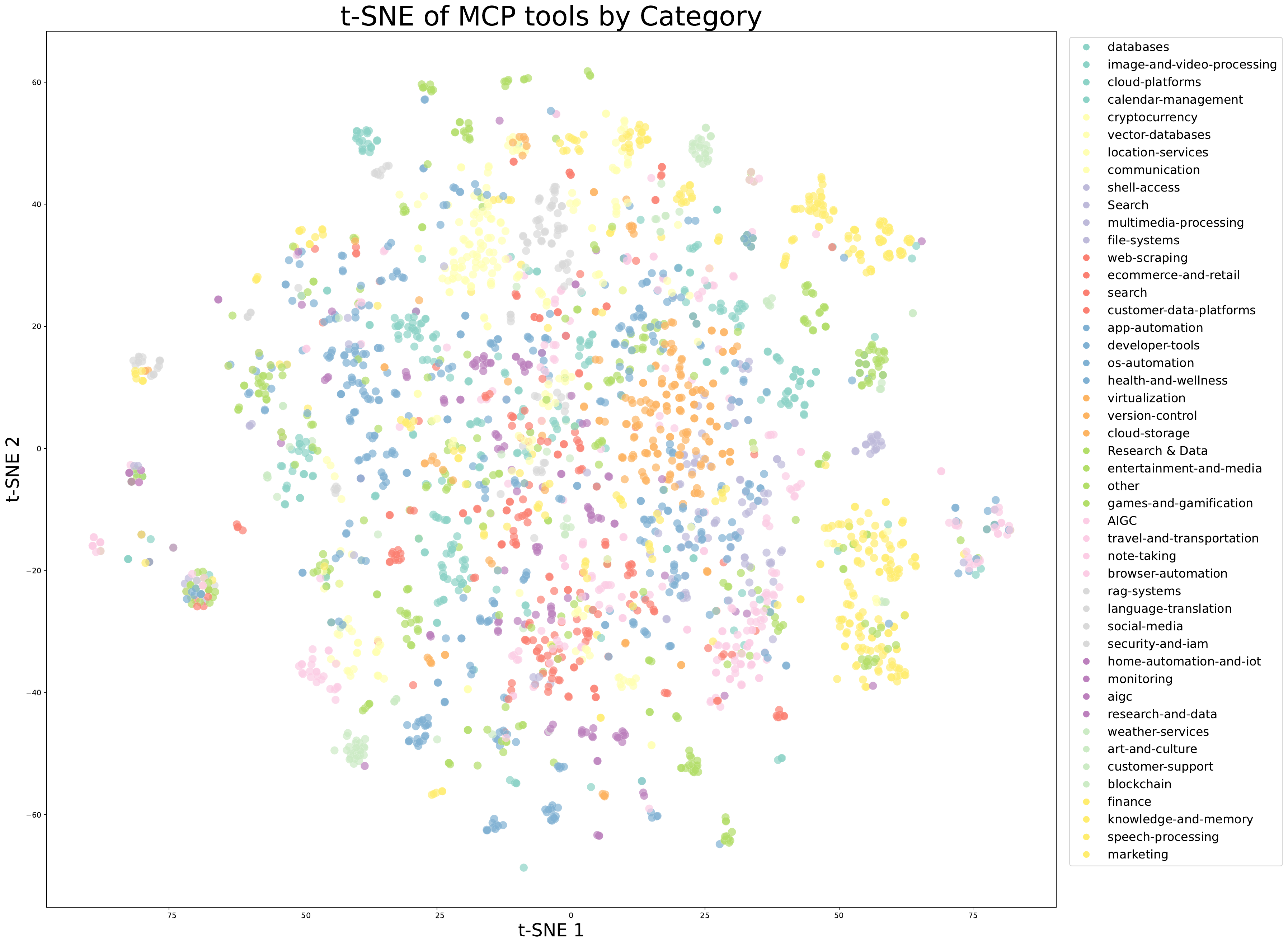}
        \caption{t-SNE visualization of real MCP tools, colored by their original source categories}
        \label{fig:tsne_mcp}
    \end{subfigure}
    \hfill
    \begin{subfigure}[b]{0.48\textwidth}
        \centering
        \includegraphics[width=\textwidth]{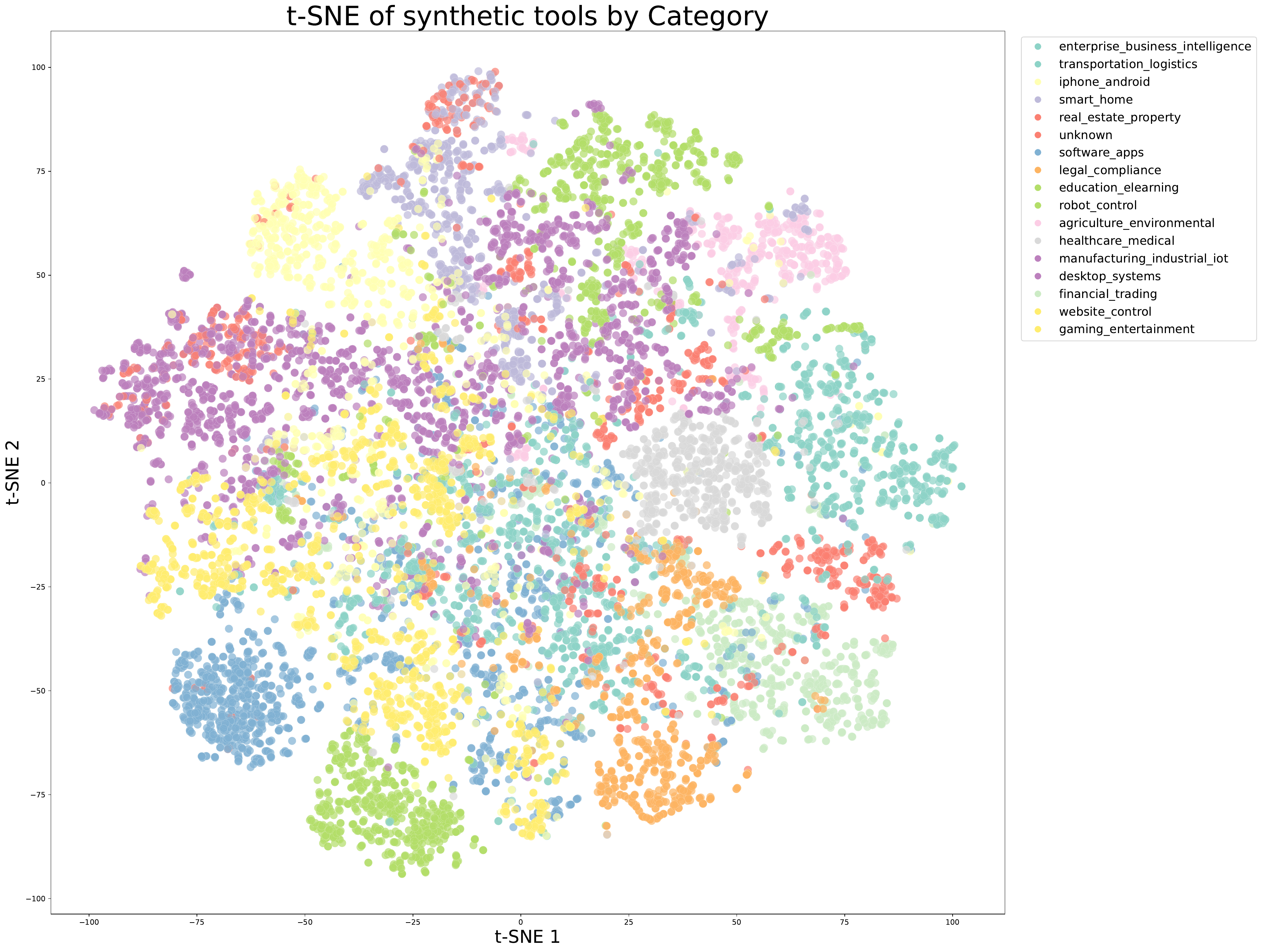}
        \caption{t-SNE visualization of synthetic tools, colored by pre-defined domain categories}
        \label{fig:tsne_synthetic}
    \end{subfigure}
    \caption{t-SNE visualizations of tool embeddings. (a) Real-world MCP tools exhibit natural clustering based on their original source categories. (b) Synthetic tools are organized into pre-defined domain categories, providing systematic coverage of the tool space. Together, they ensure comprehensive representation across different tool functionalities.}
    \label{fig:tsne_tools}
\end{figure}

There are three stages in our data synthesis pipeline, depicted in Fig.~\ref{fig:tool_synth}.
\begin{itemize}
    \item \textit{Tool spec generation}: we first construct a large repository of tool specs from both real-world tools
    and LLM-synthetic tools;
    \item \textit{Agent and task generation}: for each tool-set sampled from the tool repository, we generate an agent to use the toolset and some corresponding tasks;
    \item \textit{Trajectory generation}: for each agent and task, we generate trajectories where the agent finishes the task by invoking tools.
\end{itemize}

\paragraph{Domain Evolution and Tool Generation.} We construct a comprehensive tool repository through two complementary approaches. First, we directly fetch 3000+ real MCP (Model Context Protocol) tools from GitHub repositories, leveraging existing high-quality tool specs. Second, we systematically evolve~\citep{xu2025wizardlmempoweringlargepretrained} synthetic tools through a hierarchical domain generation process: we begin with key categories (e.g., financial trading, software applications, robot control), then evolve multiple specific application domains within each category. Specialized tools are then synthesized for each domain, with clear interfaces, descriptions, and operational semantics. This evolution process produces over 20,000 synthetic tools. Figure~\ref{fig:tsne_tools} visualizes the diversity of our tool collection through t-SNE embeddings, demonstrating that both MCP and synthetic tools cover complementary regions of the tool space.

\paragraph{Agent Diversification.} We generate thousands of distinct agents by synthesizing various system prompts and equipping them with different combinations of tools from our repository. This creates a diverse population of agents with varied capabilities, areas of expertise, and behavioral patterns, ensuring a broad coverage of potential use cases.

\paragraph{Rubric-Based Task Generation.} For each agent configuration, we generate tasks that range from simple to complex operations. Each task is paired with an explicit rubric that specifies success criteria, expected tool-use patterns, and evaluation checkpoints. This rubric-based approach ensures a consistent and objective evaluation of agent performance.

\paragraph{Multi-turn Trajectory Generation.} We simulate realistic tool-use scenarios through several components:
\begin{itemize}
    \item \textit{User Simulation}: LLM-generated user personas with distinct communication styles and preferences engage in multi-turn dialogues with agents, creating naturalistic interaction patterns.
    \item \textit{Tool Execution Environment}: A sophisticated tool simulator (functionally equivalent to a world model) executes tool calls and provides realistic feedback. The simulator maintains and updates state after each tool execution, enabling complex multi-step interactions with persistent effects. It introduces controlled stochasticity to produce varied outcomes including successes, partial failures, and edge cases.
\end{itemize}

\paragraph{Quality Evaluation and Filtering.} An LLM-based judge evaluates each trajectory against the task rubrics. Only trajectories that meet the success criteria are retained for training, ensuring high-quality data while allowing natural variation in task-completion strategies.

\paragraph{Hybrid Approach with Real Execution Environments.} While simulation provides scalability, we acknowledge the inherent limitation of simulation fidelity. To address this, we complement our simulated environments with real execution sandboxes for scenarios where authenticity is crucial, particularly in coding and software engineering tasks. These real sandboxes execute actual code, interact with genuine development environments, and provide ground-truth feedback through objective metrics such as test suite pass rates. This combination ensures that our models learn from both the diversity of simulated scenarios and the authenticity of real executions, significantly strengthening practical agent capabilities.

By leveraging this hybrid pipeline that combines scalable simulation with targeted real-world execution, we generate diverse, high-quality tool-use demonstrations that balance coverage and authenticity. The scale and automation of our synthetic data generation, coupled with the grounding provided by real execution environments, effectively implements large-scale rejection sampling~\cite{huang2022large,zelikman2022star} through our quality filtering process. This high-quality synthetic data, when used for supervised fine-tuning, has demonstrated significant improvements in the model's tool-use capabilities across a wide range of real-world applications.

\subsection{Reinforcement Learning}
Reinforcement learning (RL) is believed to have better token efficiency and generalization than SFT.
Based on the work of K1.5~\citep{team2025kimi}, we continue to scale RL in both task diversity and training FLOPs in K2. To support this, we develop a Gym-like extensible framework that facilitates RL across a wide range of scenarios. We extend the framework with a large number of tasks with verifiable rewards. For tasks that rely on subjective preferences, such as creative writing and open-ended question answering, we introduce a self-critic reward in which the model performs pairwise comparisons to judge its own outputs. This approach allows tasks from various domains to all benefit from the RL paradigm. 

\subsubsection{Verifiable Rewards Gym}

\paragraph{Math, STEM and Logical Tasks}
For math, stem and logical reasoning domains, our RL data preparation follows two key principles, \emph{diverse coverage} and \emph{moderate difficulty}.

\emph{Diverse Coverage.} 
For math and stem tasks, we collect high-quality QA pairs using a combination of expert annotations, internal QA extraction pipelines, and open datasets~\citep{li2024numinamath,moshkov2025aimo}. During the collection process, we leverage a tagging system to deliberately increase coverage of under-covered domains. For logical tasks, our dataset comprises a variety of formats, including structured data tasks (e.g., multi-hop tabular reasoning, cross-table aggregation) and logic puzzles (e.g., the 24-game, Sudoku, riddles, cryptarithms, and Morse-code decoding).

\emph{Moderate Difficulty.} The RL prompt-set should be neither too easy nor too hard, both of which may produce little signal and reduce learning efficiency. We assess the difficulty of each problem using the SFT model's pass@k accuracy and select only problems with moderate difficulty.

\paragraph{Complex Instruction Following}
Effective instruction following requires not only understanding explicit constraints but also navigating implicit requirements, handling edge cases, and maintaining consistency over extended dialogues. We address these challenges through a hybrid verification framework that combines automated verification with adversarial detection, coupled with a scalable curriculum generation pipeline.
Our approach employs a dual-path system to ensure both precision and robustness:

\noindent\textit{Hybrid Rule Verification.} We implement two verification mechanisms: (1) deterministic evaluation via code interpreters for instructions with verifiable outputs (e.g., length, style constraints), and (2) LLM-as-judge evaluation for instructions requiring nuanced understanding of constraints. To address potential adversarial behaviors where models might claim instruction fulfillment without actual compliance, we incorporate an additional hack-check layer that specifically detects such deceptive claims.

\noindent\textit{Multi-Source Instruction Generation.} To construct our training data, we employ three distinct generation strategies to ensure comprehensive coverage: (1) expert-crafted complex conditional prompts and rubrics developed by our data team (2) agentic instruction augmentation inspired by AutoIF~\citep{dong2024selfplayexecution}, and (3) a fine-tuned model specialized for generating additional instructions that probe specific failure modes or edge cases. This multipronged approach ensures both breadth and depth in instruction coverage.

\paragraph{Faithfulness}

Faithfulness is essential for an agentic model operating in scenarios such as multi-turn tool use, self-generated reasoning chains, and open-environment interactions.
Inspired by the evaluation framework from FACTS Grounding~\citep{jacovi2025factsgroundingleaderboardbenchmarking}, we train a sentence-level faithfulness judge model to perform automated verification. The judge is effective in detecting sentences that make a factual claim without supporting evidence in context. It serves as a reward model to enhance overall faithfulness performance. 

\paragraph{Coding \& Software Engineering}

To enhance our capability in tackling competition-level programming problems, we gather problems and their judges from both open-source datasets~\citep{huang2025opencoderopencookbooktoptier,xu2025kodcodediversechallengingverifiable} and synthetic sources. To ensure the diversity of the synthetic data and the correctness of reward signals, we incorporate high-quality human-written unit tests retrieved from pre-training data. 

For software engineering tasks, we collect a vast amount of pull requests and issues from GitHub to build software development environment that consists of user prompts/issues and executable unit tests.
This environment was built on a robust sandbox infrastructure, powered by Kubernetes for scalability and security. It supports over 10,000 concurrent sandbox instances with stable performance, making it ideal for both competitive coding and software engineering tasks.

\paragraph{Safety}
Our work to enhance the safety begins with a human-curated set of seed prompts, manually crafted to encompass prevalent risk categories such as violence, fraud, and discrimination.

To simulate sophisticated jailbreak attempts~(e.g., role-playing, literary narratives, and academic discourse), we employ an automated prompt evolution pipeline with three key components:
\begin{itemize}
    \item \textbf{Attack Model}: Iteratively generates adversarial prompts designed to elicit unsafe responses from the target LLM.
    \item \textbf{Target Model}: Produces responses to these prompts, simulating potential vulnerabilities.
    \item \textbf{Judge Model}: Evaluates the interaction to determine if the adversarial prompt successfully bypasses safety mechanisms.
\end{itemize}

Each interaction is assessed using a task-specific rubric, enabling the judge model to provide a binary success/failure label.

\subsubsection{Beyond Verification: Self-Critique Rubric Reward}

To extend model alignment beyond tasks with verifiable reward, we introduce a framework for general reinforcement learning from self-critic feedbacks. This approach is designed to align LLMs with nuanced human preferences, including helpfulness, creativity, depth of reasoning, factuality, and safety, by extending the capabilities learned from verifiable scenarios to a broader range of subjective tasks.  The framework operates using a \textit{Self-Critique Rubric Reward} mechanism, where the model evaluates its own outputs to generate preference signals. 
To bootstrap K2 as a competent judge, we curated a mixture of open-source and in-house preference datasets and initialize its critic capability in the SFT stage.

\paragraph{Self-Critiqued Policy Optimization}
In the first core process of the learning loop, the K2 actor generates responses for general prompts that cover a wide range of use cases. The K2 critic then ranks all results by performing pairwise evaluations against a combination of rubrics, which incorporates both \textit{core rubrics} (Appendix.~\ref{sec:k2-critic-creteria}), which represent the fundamental values of our AI assistant that Kimi cherish, prescriptive rubrics (Appendix.~\ref{sec:k2-must-not-creteria}) that aim to eliminate reward hacking, and \textit{human-annotated rubrics} crafted by our data team for specific instructional contexts. Although certain rubrics can be designated as mandatory, K2 retains the flexibility to weigh them against its internal priors. This capacity enables a dynamic and continuous alignment with its evolving on-policy behavior, ensuring that the model's responses remain coherent with its core identity while adapting to specific instructions.

\paragraph{Closed-Loop Critic Refinement and Alignment}
During RL training, the critic model is refined using verifiable signals. On-policy rollouts generated from verifiable-reward prompts are used to continuously update the critic, a crucial step that distills objective performance signals from RLVR directly into its evaluation model. This transfer learning process grounds its more subjective judgments in verifiable data, allowing the performance gains from verifiable tasks to enhance the critic's judgment on complex tasks that lack explicit reward signals. 
This closed-loop process ensures that the critic continuously recalibrates its evaluation standards in lockstep with the policy's evolution. By grounding subjective evaluation in verifiable data, the framework enables robust and scalable alignment with complex, non-verifiable human objectives. 

Consequently, this holistic alignment yields comprehensive performance improvements across a wide spectrum of domains, including user intent understanding, creative writing, complex reasoning, and nuanced language comprehension.

\subsubsection{RL Algorithm}

We adopt the policy optimization algorithm introduced in K1.5~\citep{team2025kimi} as the foundation for K2. 
For each problem $x$, we sample $K$ responses $\{y_1,\dots,y_k\}$ from the previous policy $\pi_{\mathrm{old}}$, 
and optimize the model $\pi_\theta$ with respect to the following objective: 
\begin{align*}
L_{\mathrm{RL}}(\theta) = \mathbb{E}_{x \sim\mathcal{D}}\left[ \frac{1}{K} \sum_{i=1}^K \left[  \left( r(x, y_i) - \bar{r}(x)- \tau \log \frac{\pi_\theta(y_i | x)}{{\pi}_{\mathrm{old}}(y_i | x)} \right)^2 \right]\right] \, ,
\label{eq:rl-objective}
\end{align*}
where $\bar{r}(x) = \frac{1}{k}\sum_{i=1}^k r(x, y_i)$ is the mean rewards of the sampled responses, $\tau>0$ is a regularization parameter that promotes stable learning. 
As in SFT, we employ the Muon optimizer~\citep{jordan2024muon} to minimize this objective. 
As we scale RL training to encompass a broader range of tasks in K2, a primary challenge is achieving consistent performance improvements across all domains. To address this, we introduce several additions to the RL algorithm.

\paragraph{Budget Control}
It has been widely observed that RL often results in a substantial increase in the length of model-generated responses~\citep{team2025kimi,guo2025deepseek}. While longer responses can enable the model to utilize additional test-time compute for improved performance on complex reasoning tasks, the benefits often do not justify its inference cost in non-reasoning domains. 
To encourage the model to properly distribute inference budget, we enforce a per-sample \emph{maximum token budget} throughout RL training, where the budget is determined based on the type of task.  
 Responses that exceed this token budget are truncated and assigned a penalty, which incentivizes the model to generate solutions within the specified limit. Empirically, this approach significantly enhances the model's token efficiency, encouraging concise yet effective solutions across all domains.

\paragraph{PTX Loss} 

To prevent the potential forgetting of valuable, high-quality data during joint RL training, we curate a dataset comprising hand-selected, high-quality samples and integrate it into the RL objective through an auxiliary PTX loss~\citep{ouyang2022training}.
This strategy not only leverages the advantages of high-quality data, but also mitigates the risk of overfitting to the limited set of tasks explicitly present in the training regime. 
This augmentation substantially improves the model's generalization across a broader range of domains.

\paragraph{Temperature Decay}

For tasks such as creative writing and complex reasoning, we find that promoting exploration via a high sampling temperature during the initial stages of training is crucial. 
A high temperature allow the model to generate diverse and innovative responses, thereby facilitating the discovery of effective strategies and reducing the risk of premature convergence to suboptimal solutions. 
However, retaining a high temperature in the later stages of training or during evaluation can be detrimental, as it introduces excessive randomness and compromises the reliability and consistency of the model's outputs. 
To address this, we employ a temperature decay schedule, to shift from exploration to exploitation throughout the training. This strategy ensures that the model leverages exploration when it is most beneficial, while ultimately converge on stable and high-quality outputs.

\subsection{RL Infrastructure}

\subsubsection{Colocated Architecture}
Similar to K1.5~\citep{team2025kimi}, we adopt a hybrid colocated architecture for our synchronized RL training, where the training and inference engines live on the same workers. When one engine is actively working, the other engine releases or offloads its GPU resources to accommodate. In each iteration of RL training, a centralized controller first calls the inference engine to generate new data for training. It then notifies the training engine to train on the new data, and send updated parameters to the inference engine for the next iteration. 

Each engine is heavily optimized for throughput. In addition, as the model scales to the size of K2, the latency of engine switching and failure recovery becomes significant. We present our system
design considerations in these aspects.

\subsubsection{Efficient Engine Switching}

During rollout, the parameters of the training engine are offloaded to DRAM. Bringing up the training engine is therefore a simple step of H2D transmission. 
However, bringing up the inference engine is a bigger challenge, as it must obtain updated parameters
from the training engine with a different
sharding paradigm.

\begin{figure}
    \centering
    \includegraphics[width=0.5\linewidth]{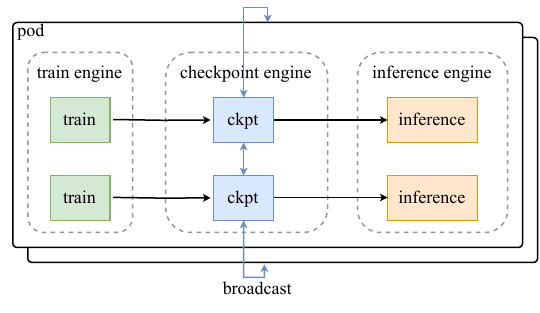}
    \caption{Parameter update utilizing a checkpoint engine}
    \label{fig:RL-weight}
\end{figure}

Given the scale of K2 and the vast number of devices involved, using a network file system for resharding and broadcasting parameters is impractical. The aggregate bandwidth required to keep overhead low reaches several petabytes per second. To address this challenge, we developed a distributed checkpoint engine co-located on training nodes to manage parameter states. To perform a parameter update, each checkpoint engine worker obtains a local copy of parameters from the training engine, then broadcasts the full parameter set across all checkpoint engine workers. Subsequently, the inference engine retrieves only the parameter shard it requires from the checkpoint engine. This process is illustrated in Figure~\ref{fig:RL-weight}. To enable this for a 1T model, updates are performed parameter-by-parameter in a pipelined manner, minimizing memory footprint (see Appendix~\ref{sec:k2-rl-update-appendix}). 

We opt to broadcast the full parameter set across the entire cluster, regardless of the specific sharding schemes on each inference worker. While this transfers several times more data than a theoretically optimal approach, it offers a simpler system design that is less intrusive to the training and inference engines. We chose to trade off this minor overhead to fully decouple the training engine and the inference engine, significantly simplifying maintenance and testing. 

Notably, this approach outperforms the transfer-what-you-need method due to reduced synchronization overhead and higher network bandwidth utilization.
Our system can complete a full parameter update for Kimi K2 with less than 30 seconds, a negligible duration for a typical RL training iteration. The source code for the checkpoint engine is available on Github\footnote{\url{https://github.com/MoonshotAI/checkpoint-engine}}.

\subsubsection{Efficient System Startup}

As large-scale training is prone to system failure, optimizing the startup time is crucial for models as large as Kimi K2.

To start the training engine, we let each training worker selectively read part or none of the parameters from disk, and broadcast necessary parameters to its peers. The design goal is to ensure all workers collectively read the checkpoint only once, minimizing expensive disk IO. 

As the inference engines are independent replicas, we would like to avoid introducing extra synchronization barriers between them. Therefore, we opt to reuse checkpoint engine for startup: we let checkpoint engine collectively read the checkpoint from disk, similar to how the training engine starts. Then it updates the state of the uninitialized inference engine, using the approach introduced in the previous section. By leveraging the dedicated checkpoint engine, the system also becomes robust to single-point failures, because an inference replica can restart without communicating with other replicas.

\subsubsection{Agentic Rollout}
Our RL infrastructure supports the training of long-horizon, multi-turn agentic tasks. During rollout, these tasks present distinct challenges, such as complex environmental interactions and prolonged rollout durations. Here we introduce a few optimizations to alleviate these issues.

Due to the diversity of environments, certain interactions may be blocked on waiting for environment feedback (e.g., a virtual machine or a code interpreter), leaving the GPUs idle. We employ two strategies to maximize GPU utilization: (i) we deploy heavy environments as dedicated services that can scale up more easily; (ii) we employ a large number of concurrent rollouts to amortize the latency induced by certain expensive interactions.

Another challenge in agentic rollout is that individual rollout trajectories can be extremely long. To prevent long-tail trajectories from blocking the entire rollout process, we employ the partial rollout~\citep{team2025kimi} technique. This strategy allows long-tail unfinished tasks to be paused, and resumed in the next RL iteration.

To improve research efficiency, we also design a unified interface inspired by the OpenAI Gym framework~\citep{openaigym2016} to streamline the integration of new environments. We hope to scale our RL infrastructure to more diverse interactive environments in the future.

%% file: 4-results.tex
\section{Evaluations}

This section begins with the post-training evaluation of Kimi-K2-Instruct, followed by a brief overview of the capabilities of Kimi-K2-Base. We conclude with a comprehensive safety evaluation.

\subsection{Post-training Evaluations}

\subsubsection{Evaluation Settings}

\paragraph{Benchmarks}
We assess Kimi-K2-Instruct across different areas.
For coding, we adopt LiveCodeBench v6~\citep{jain2024livecodebench}(questions from August 2024 to May 2025), OJBench~\citep{wang2025ojbenchcompetitionlevelcode}, MultiPL-E~\citep{10103177}, SWE-bench Verified~\citep{jimenez2024swebench,yang2025swesmith}, TerminalBench~\citep{tbench_2025}, Multi-SWE-bench~\citep{zan2025multi}, SWE-Lancer~\citep{miserendino2025swelancer}, PaperBench~\citep{starace2025paperbench}, and Aider-Polyglot~\citep{aider}. 
For tool use tasks, we evaluate performance on $\tau^2$-Bench~\citep{barres2025tau2} and AceBench~\citep{chen2025acebench}, which emphasize multi-turn tool-calling capabilities.
In reasoning, we include a wide range of mathematical, science and logical tasks: AIME 2024/2025, MATH-500, HMMT 2025, CNMO 2024, PolyMath-en, ZebraLogic~\citep{lin2025zebralogicscalinglimitsllms}, AutoLogi~\citep{zhu2025autologiautomatedgenerationlogic}, GPQA-Diamond~\citep{rein2024gpqa}, SuperGPQA~\citep{du2025supergpqa}, and Humanity's Last Exam (Text-Only)~\citep{phan2025humanitysexam}.
We benchmark the long-context capabilities on:  MRCR\footnote{\url{https://huggingface.co/datasets/openai/mrcr}} for long-context retrieval, and  DROP~\citep{DBLP:journals/corr/abs-1903-00161}, FRAMES~\citep{krishna2025factfetchreasonunified}  and  LongBench v2~\citep{bai2025longbenchv2deeperunderstanding} for  long-context reasoning.
For factuality, we evaluate FACTS Grounding~\citep{jacovi2025factsgroundingleaderboardbenchmarking}, the Vectara Hallucination Leaderboard~\citep{hhem-2.1-open}, and FaithJudge~\citep{tamber2025benchmarking}.
Finally, general capabilities are assessed using MMLU~\citep{hendrycks2021measuringmassivemultitasklanguage}, MMLU-Redux~\citep{gema2024we}, MMLU-Pro~\citep{wang2024mmluprorobustchallengingmultitask}, IFEval~\citep{Zhou2023InstructionFollowingEF}, Multi-Challenge~\citep{sirdeshmukh2025multichallengerealisticmultiturnconversation}, SimpleQA~\citep{wei2024measuring}, and LiveBench~\citep{livebench} (as of 2024-11-25).

\paragraph{Baselines}{
We benchmark against both open-source and proprietary frontier models, ensuring every candidate is evaluated under its non-thinking configuration to eliminate additional gains from test-time compute.
Open-source baselines: DeepSeek-V3-0324 and Qwen3-235B-A22B, with the latter run in the vendor-recommended no-thinking regime.
Proprietary baselines: Claude Sonnet 4, Claude Opus 4, GPT-4.1, and Gemini 2.5 Flash Preview (2025-05-20). Each invoked in its respective non-thinking mode via official APIs under unified temperature and top-p settings.

\textbf{Evaluation Configurations} All runs query models in their non-thinking mode. Output token length is capped at 8192 tokens everywhere except SWE-bench Verified (Agentless), which is raised to 16384. 
For benchmarks with high per-question variance, we adopt repeated sampling $k$ times and average the results to obtain stable scores, denoted as Avg@k.
For long-context tasks, we set the context window size to 128K tokens during evaluation, truncating any input that exceeds this limit to fit within the window.
SWE-bench Verified is evaluated in two modes: Agentless Coding via Single Patch without Test (Acc) and Agentic Coding via bash/editor tools under both Single Attempt (Acc) and Multiple Attempts (Acc) using best-of-N selection with an internal verifier; SWE-bench Multilingual is tested only in the single-attempt agentic setting. Some data points have been omitted due to prohibitively expensive evaluation costs.
}

\input{tables/instruct_eval}

\subsubsection{Evaluation Results}

A comprehensive evaluation results of Kimi-K2-Instruct is shown in Table~\ref{tab:instruct_eval}, with detailed explanation provided in the Appendix~\ref{sec:k2-post-train-eval-details}. Below, we highlight key results across four core domains:

\paragraph{Agentic and Competitive Coding}
Kimi-K2-Instruct demonstrates state-of-the-art open-source performance on real-world SWE tasks. It outperforms most baselines on SWE-bench Verified (65.8\%, 71.6\% with multiple attemps), SWE-bench Multilingual (47.3\%), and SWE-lancer (39.1\%), significantly closing the gap with Claude 4 Opus and Sonnet. On competitive coding benchmarks (e.g., LiveCodeBench v6 53.7\%, OJBench 27.1\%), it also leads among all models, highlighting its practical coding proficiency across difficulty levels.

\paragraph{Agentic Tool Use}
On multi-turn tool-use benchmarks, Kimi-K2-Instruct sets a new standard. It achieves 66.1 Pass@1 on $\tau^2$-Bench and 76.5 on ACEBench, substantially outperforming all baselines. These results affirm its strength in grounded, controlled, and agent-driven tool orchestration across domains.

\paragraph{General Capabilities}
Kimi-K2-Instruct exhibits strong, balanced performance across general knowledge, math, instruction following, and long-context tasks. It surpasses open-source peers on SimpleQA (31.0\%), MMLU (89.5\%) and MMLU-Redux (92.7\%), and leads all models on instruction benchmarks (IFEval: 89.8\%, Multi-Challenge: 54.1\%). In math and STEM, it achieves top-tier scores (AIME 2024: 69.6\%, GPQA-Diamond: 75.1\%), and remains competitive on long-context factuality and retrieval (DROP: 93.5\%, MRCR: 55.0\%). These results position Kimi-K2-Instruct as a well-rounded and capable generalist across both short- and long-context settings.

\paragraph{Open-Ended Evaluation}
On the LMSYS Arena leaderboard (July 17, 2025), Kimi-K2-Instruct ranks as the top-1 open-source model and 5th overall based on over 3,000 user votes. This real-world preference signal—across diverse, blind prompts—underscores Kimi-K2's strengths in generating high-quality responses on open-ended tasks.

\subsection{Pre-training Evaluations}

\subsubsection{Evaluation Settings}

\paragraph{Benchmarks}
We evaluate Kimi-K2-Base across diverse capability areas.
For general capabilities, we assess on MMLU~\citep{hendrycks2021measuringmassivemultitasklanguage}, MMLU-Pro~\citep{wang2024mmluprorobustchallengingmultitask}, MMLU-Redux~\citep{gema2024we}, BBH~\citep{suzgun2022challengingbigbenchtaskschainofthought}, TriviaQA~\citep{joshi2017triviaqalargescaledistantly}, SuperGPQA~\citep{du2025supergpqa}, SimpleQA~\citep{wei2024measuring}, HellaSwag~\citep{zellers2019hellaswag}, AGIEval~\citep{zhong2023agieval}, GPQA-Diamond~\citep{rein2024gpqa}, ARC-Challenge~\citep{clark2018think}, and WinoGrande~\citep{sakaguchi2021winogrande}.
For coding capabilities, we employ EvalPlus~\citep{liu2023your} (averaging HumanEval~\citep{chen2021codex}, MBPP~\citep{austin2021programsynthesislargelanguage}, HumanEval+, and MBPP+), LiveCodeBench v6~\citep{jain2024livecodebench}, and CRUXEval~\citep{gu2024cruxeval}.
For mathematical reasoning, we utilize GSM8K~\citep{cobbe2021trainingverifierssolvemath}, GSM8K-Platinum~\citep{vendrow2025large}, MATH~\citep{hendrycks2021measuringmathematicalproblemsolving}, and CMATH~\citep{wei2023cmathlanguagemodelpass}.
For Chinese language capabilities, we evaluate on C-Eval~\citep{huang2023cevalmultilevelmultidisciplinechinese}, CMMLU~\citep{li2024cmmlumeasuringmassivemultitask}, and CSimpleQA~\citep{he2411chinese}.

\paragraph{Baselines}
We benchmark against leading open-source foundation models: DeepSeek-V3-Base~\citep{deepseekai2024deepseekv3technicalreport}, Qwen2.5-72B-Base~\citep{qwen2025qwen25technicalreport} (Note that Qwen3-235B-A22B-Base is not open-sourced, and the largest open-sourced base model in the Qwen series is Qwen2.5-72B-Base), and Llama 4-Maverick~\citep{metaLlamaHerd} (Llama 4-Behemoth is also not open-sourced). All models are evaluated under identical configurations to ensure fair comparison.

\paragraph{Evaluation Configurations}

We employ perplexity-based evaluation for MMLU, MMLU-Redux, GPQA-Diamond, HellaSwag, ARC-Challenge, C-Eval, and CMMLU. Generation-based evaluation is used for MMLU-Pro, SuperGPQA, TriviaQA, BBH, CSimpleQA, MATH, CMATH, GSM8K, GSM8K-Platinum, CRUXEval, LiveCodeBench, and EvalPlus. To mitigate the high variance inherent to GPQA-Diamond, we report the mean score across eight independent runs. All evaluations are conducted using our internal framework derived from LM-Harness-Evaluation~\citep{biderman2024lessons}, ensuring consistent settings across all models.

\subsubsection{Evaluation Results}

Table~\ref{tab:pretrain-eval} presents a comprehensive comparison of Kimi-K2-Base against leading open-source foundation models across diverse evaluation benchmarks. The results demonstrate that Kimi-K2-Base achieves state-of-the-art performance across the majority of evaluated tasks, establishing it as a leading foundation model in the open-source landscape.

\paragraph{General Language Understanding}
Kimi-K2-Base achieves state-of-the-art performance on 10 out of 12 English language benchmarks. Notable results include MMLU (87.79\%), MMLU-Pro (69.17\%), MMLU-Redux (90.17\%), SuperGPQA (44.67\%), and SimpleQA (35.25\%), significantly outperforming all baselines.

\paragraph{Coding Capabilities}
On coding benchmarks, Kimi-K2-Base sets new standards with leading performance across all metrics. It achieves 74.00\% on CRUXEval-I-cot, 83.50\% on CRUXEval-O-cot, 26.29\% on LiveCodeBench v6, and 80.33\% on EvalPlus, demonstrating superior code generation and comprehension abilities, particularly in scenarios requiring step-by-step reasoning.

\paragraph{Mathematical Reasoning}
Kimi-K2-Base exhibits exceptional mathematical capabilities, leading on three out of four benchmarks: MATH (70.22\%), GSM8K (92.12\%), and GSM8K-Platinum (94.21\%). It maintains competitive performance on CMATH (90.26\%), narrowly behind DeepSeek-V3-Base (90.53\%). These results highlight the model's robust mathematical problem-solving abilities across varying difficulty levels.

\paragraph{Chinese Language Understanding}
The model demonstrates superior multilingual capabilities, achieving state-of-the-art results across all Chinese language benchmarks: C-Eval (92.50\%), CMMLU (90.90\%), and CSimpleQA (77.57\%). These results establish Kimi-K2-Base as a leading model for Chinese language understanding while maintaining strong performance across other languages.

\input{tables/base_eval}

\subsection{Safety Evaluation}

\subsubsection{Experiment Settings}

We conducted red-teaming evaluations on Kimi K2 compare with other open-source LLMs. The evaluation covered a range of attack scenarios—including harmful content, privacy content, and security content, as well as different attack strategies such as prompt injection and iterative jailbreak.

We choose \emph{Promptfoo}\footnote{\url{https://github.com/promptfoo/promptfoo}} to generate adversarial prompts and analyze the responses. By this way, we can evaluate model in a scalable ways.

\textbf{Model Selection} We compare Kimi K2 with three other open-source LLMs: DeepSeek-V3, DeepSeek-R1, and Qwen3.

\textbf{Promptfoo Settings} Table~\ref{tab:safety-stategy} lists plugins and strategies evaluated, with each plugin paired with all strategies to assess their performance.

\renewcommand{\arraystretch}{1.2}
\begin{table}[h]
\centering
\footnotesize
\caption{Enabled Plugins and Strategies}
\label{tab:safety-stategy}
\setlength{\tabcolsep}{1.9pt}
\begin{tabular}{|c|c|p{12cm}|}    
\hline
\multirow{5}{*}{Plugin} & Harmful & Graphic Content, Harassment and Bullying, Hate Speech, Insults, Profanity, Radicalization, Self Harm, Sexual Content, ToxicChat \\ \cline{2-3}
                      & Criminal & Chemical\&Biological Weapons, Child Exploitation, Copyright Violations, Cybercrime, Illegal Activities, Illegal Drugs, Indiscriminate Weapons, Intellectual Property Violation, Non-Violent Crime, Violent Crime, Sex Crimes \\ \cline{2-3}
                      & Misinformation & Competitor Endorsement, Unsupervised Contracts, Excessive Agency, Hallucination, Misinformation and Disinformation, Specialized Advice, Unsafe Practices, Imitation, Overreliance, Political Opinions, Religious Sensitivity \\ \cline{2-3}
                      & Privacy & Privacy Violation, PII in API/Database, Direct PII Exposure, PII in Session Data, PII via Social Engineering \\ \cline{2-3}
                      & Security & ASCII Smuggling, CyberSecEval, Harmbench, Debug Access, Divergent Repetition, DoNotAnswer, Malicious Code, Pliny, Prompt Extraction, Reasoning DoS, Tool Discovery \\ \hline
Strategy              & \multicolumn{2}{p{10cm}|}{Basic, Prompt Injection, Iterative Jailbreak, Crescendo} \\ \hline
\end{tabular}
\label{tab:plugin-strategy}
\end{table}
\renewcommand{\arraystretch}{1.0}

\textbf{Test Case Count} Given the inherent non-determinism of large language model inference, single-pass outputs may exhibit variability. To account for this, we generated 3 attack prompts per plugin for each strategy.

\textbf{Prompt Language Settings} We pre-tested the language compatibility for each plugin-strategy combination. Some plugins support both English and Chinese, while others only support English. For combinations that support both, we generated 3 prompts in each language, resulting in 6 prompts per combination.

\textbf{Manual Review} We incorporated human review into the evaluation process. To minimize subjectivity problem, we conducted multiple rounds of review and assigned the same reviewer to evaluate all cases within a given test set to ensure consistency and reduce variability in judgment.

\subsubsection{Safety Evaluation Results}
Table~\ref{tab:safety-eval} presents the passing rates of different models under various plugin–strategy combinations.

\input{tables/safety_eval}

Without targeted optimization for specific evaluation scenarios, the passing rate of some complex cases (e.g., Harmful–Iterative Jailbreak) was relatively higher compared to other models.

Across different attack strategies, the models exhibited varying trends. Under the Base64 strategy, passing rates generally approached or reached 100\%, suggesting that encoding transformations had minimal impact on the models’ basic robustness. In contrast, the Crescendo strategy led to a general drop in passing rates, indicating stronger adversarial effectiveness.

In addition, complex attack strategies do not always outperform basic prompts. Some originally adversarial prompts may lose their intended meaning after multiple rounds of transformation, rendering the resulting model outputs less meaningful.

\textbf{Automated Red-teaming Limitations }Due to the involvement of human review, the evaluation results inevitably contain a degree of subjectivity. Additionally, certain plugin types involve API misuse or external tool invocation, which are more suitable for evaluating agent models with tool-calling capabilities. In the context of base LLMs, such tests may have limited relevance.

%% file: tables/instruct_eval.tex
\newcolumntype{P}[1]{>{\centering\arraybackslash}p{#1}}

\begin{table}[htbp] % 使用 [htbp] 提供更灵活的浮动位置
\centering
\footnotesize % 使用 footnotesize，比 scriptsize 稍大，更易读
\setlength{\tabcolsep}{4pt} % 减少列间距以节省空间

\caption{Performance comparison of Kimi-K2-Instruct against leading open-source and proprietary models across diverse tasks. \textbf{Bold} denotes the global SOTA; \underline{\textbf{underlined bold}} indicates the best open-source result. Data points marked with * are taken directly from the model's technical report or blog.}
\label{tab:instruct_eval} % 添加一个标签，方便在文中引用
\begin{tabular}{@{}l *{7}{P{1.4cm}}@{}} % *{7}{P{1.6cm}} 表示重复 P{1.6cm} 7次
\toprule
& \multicolumn{3}{c}{\textbf{Open Source}} & \multicolumn{4}{c}{\textbf{Proprietary}} \\ % 调整了 & 数量
\cmidrule(lr){2-4} \cmidrule(lr){5-8} % 调整了 cmidrule 的列范围
\textbf{Benchmark} & \textbf{Kimi-K2-Instruct} & \textbf{DeepSeek-V3-0324} & \textbf{Qwen3-235B-A22B} & \textbf{Claude Sonnet 4} & \textbf{Claude Opus 4} & \textbf{GPT-4.1} & \textbf{Gemini 2.5 Flash} \\
\midrule
\multicolumn{8}{@{}l}{\textbf{Coding Tasks}} \\ % 调整为 8 列
\midrule
LiveCodeBench~v6 (Pass@1) & \textbf{53.7} & 46.9 & 37.0 & 48.5 & 47.4 & 44.7 & 44.7 \\
OJBench (Pass@1) & \textbf{27.1} & 24.0 & 11.3 & 15.3 & 19.6 & 19.5 & 19.5 \\
MultiPL-E (Pass@1) & \textbf{\underline{85.7}} & 83.1 & 78.2 & 88.6 & \textbf{89.6} & 86.7 & 85.6 \\
\makecell[l]{SWE-bench~Verified\\\textit{Agentless-Single-Patch} (Pass@1)} & \textbf{\underline{51.8}} & 36.6 & 39.4 & 50.2 & \textbf{53.0} & 40.8 & 32.6 \\
\makecell[l]{SWE-bench~Verified\\\textit{Agentic-Single-Attempt} (Pass@1)} & \textbf{\underline{65.8}} & 38.8 & 34.4 & \textbf{72.7*} & 72.5* & 54.6 & --- \\
\makecell[l]{SWE-bench~Verified\\\textit{Agentic-Multi-Attempt} (Pass@1)} & \textbf{\underline{71.6}} & --- & --- & \textbf{80.2*} & 79.4* & --- & --- \\
\makecell[l]{SWE-bench Multilingual (Pass@1)} & \textbf{\underline{47.3}} & 25.8 & 20.9 & \textbf{51.0} & --- & 31.5 & --- \\
\makecell[l]{Multi-SWE-bench (Pass@1)} & \textbf{\underline{18.3}} & 8.0 & 9.0 & \textbf{29.2}  & --- & 11.7 & 14.0 \\
\makecell[l]{SWE-Lancer (Pass@1)} & \textbf{\underline{39.1}} & 30.5 & 24.1 & \textbf{40.8}  & --- & 23.0 & 38.5 \\
\makecell[l]{Paper Bench~\textit{Code-Dev} (Acc.)} & \textbf{\underline{27.8}} & 12.2 & 13.2 & \textbf{43.3}  & --- & 29.9 & 5.7 \\

Terminal Bench \textit{In-House} (Acc.) & \textbf{\underline{30.0}} & --- & --- & 35.5 & \textbf{43.2} & 8.3 & --- \\
Terminal Bench \textit{Terminus} (Acc.) & \textbf{\underline{25.0}} & 16.3 & 6.6 & --- & --- & \textbf{30.3} & 16.8 \\
Aider-Polyglot (Acc.) & 60.0 & 55.1 & \textbf{\underline{61.8}} & 56.4 & \textbf{70.7} & 52.4 & 44.0 \\
\midrule
\multicolumn{8}{@{}l}{\textbf{Tool Use Tasks}} \\ % 调整为 8 列
\midrule
Tau2 retail (Avg@4) & \textbf{\underline{70.6}} & 69.1 & 57.0 & 75.0 & \textbf{81.8} & 74.8 & 64.3 \\
Tau2 airline (Avg@4) & \textbf{\underline{56.5}} & 39.0 & 26.5 & 55.5 & \textbf{60.0} & 54.5 & 42.5 \\
Tau2 telecom (Avg@4) & \textbf{65.8} & 32.5 & 22.1 & 45.2 & 57.0 & 38.6 & 16.9 \\
AceBench (Acc.) & \textbf{\underline{76.5}} & 72.7 & 70.5 & 76.2 & 75.6 & \textbf{80.1} & 74.5 \\
\midrule
\multicolumn{8}{@{}l}{\textbf{Math \& STEM Tasks}} \\ % 调整为 8 列
\midrule
AIME 2024 (Avg@64) & \textbf{69.6} & 59.4* & 40.1* & 43.4 & 48.2 & 46.5 & 61.3 \\
AIME 2025 (Avg@64) & \textbf{49.5} & 46.7 & 24.7* & 33.1* & 33.9* & 37.0 & 46.6 \\
MATH-500 (Acc.) & \textbf{97.4} & 94.0* & 91.2* & 94.0 & 94.4 & 92.4 & 95.4 \\
HMMT 2025 (Avg@32) & \textbf{38.8} & 27.5 & 11.9 & 15.9 & 15.9 & 19.4 & 34.7 \\
CNMO 2024 (Avg@16) & 74.3 & \textbf{\underline{74.7}} & 48.6 & 60.4 & 57.6 & 56.6 & \textbf{75.0} \\
PolyMath-en (Avg@4) & \textbf{65.1} & 59.5 & 51.9 & 52.8 & 49.8 & 54.0 & 49.9 \\
ZebraLogic (Acc.) & \textbf{89.0} & 84.0 & 37.7* & 79.7 & 59.3 & 58.5 & 57.9 \\
AutoLogi (Acc.) & \textbf{\underline{89.5}} & 88.9 & 83.3* & \textbf{89.8} & 86.1 & 88.2 & 84.1 \\
GPQA-Diamond (Avg@8) & \textbf{75.1} & 68.4* & 62.9* & 70.0* & 74.9* & 66.3 & 68.2 \\
SuperGPQA (Acc.) & \textbf{57.2} & 53.7 & 50.2 & 55.7 & 56.5 & 50.8 & 49.6 \\
Humanity's Last Exam (Acc.) & 4.7 & 5.2 & \textbf{\underline{5.7}} & 5.8 & \textbf{7.1} & 3.7 & 5.6 \\
\midrule
\multicolumn{8}{@{}l}{\textbf{General Tasks}} \\ % 调整为 8 列
\midrule
MMLU (EM) & \textbf{\underline{89.5}} & 89.4 & 87.0 & 91.5 & \textbf{92.9} & 90.4 & 90.1 \\
MMLU-Redux (EM) & \textbf{\underline{92.7}} & 90.5 & 89.2* & 93.6 & \textbf{94.2} & 92.4 & 90.6 \\
MMLU-Pro (EM) & 81.1 & \textbf{\underline{81.2*}} & 77.3 & 83.7 & \textbf{86.6} & 81.8 & 79.4 \\
IFEval (Prompt Strict) & \textbf{89.8} & 81.1 & 83.2* & 87.6 & 87.4 & 88.0 & 84.3 \\
Multi-Challenge (Acc.) & \textbf{54.1} & 31.4 & 34.0 & 46.8 & 49.0 & 36.4 & 39.5 \\
SimpleQA (Correct) & \textbf{\underline{31.0}} & 27.7 & 13.2 & 15.9 & 22.8 & \textbf{42.3} & 23.3 \\
Livebench (Pass@1) & \textbf{76.4} & 72.4 & 67.6 & 74.8 & 74.6 & 69.8 & 67.8 \\
\makecell[l]{Arena Hard v2.0 \\\emph{Hard Prompt} (Win rate)} & \textbf{\underline{54.5}} & 39.9 & 39.9 & 51.6 & \textbf{59.7} & 51.7 & 48.7 \\
\makecell[l]{Arena Hard v2.0 \\\emph{Creative Writing} (Win rate)} & \textbf{85.0} & 59.3 & 59.8 & 54.6 & 68.5 & 61.5 & 72.8 \\
FACTS Grounding (Adjusted) & \textbf{88.5} & 68.3 & 68.5 & 83.6 & --- & 79.2 & 86.6 \\
HHEM v2.1 (1-Hallu.) & \textbf{98.9} & 88.9 & 94.5 & 94.5 & --- & 96.7 & 97.8 \\
FaithJudge (1-Hallu.) & \textbf{\underline{92.6}} & 83.4 & 75.7 & 83.0 & --- & 91.0 & \textbf{93.2} \\
LongBench v2 (Acc.) &49.1&\textbf{\underline{51.1}}&---&52.5& --- & 54.3 & \textbf{55.5}\\
FRAMES (Acc.)  & 77.1 & \textbf{\underline{79.2}}&--- & 76.3&---&\textbf{87.4} &72.9\\
MRCR (Acc.) & \textbf{\underline{55.0}}& 50.8&---&74.4&---&66.9&\textbf{81.7}\\
DROP (Acc.) &\textbf{93.5}&91.2&84.3&92.0&---&79.1&81.7\\
\bottomrule
\\[-0.5ex]               % 让标题往下沉 0.5ex（可微调）
\end{tabular}
\end{table}

%% file: tables/base_eval.tex
\begin{table}[h]
\centering
\footnotesize
\setlength{\tabcolsep}{1.9pt}
\caption{Performance comparison of Kimi-K2-Base against leading open-source models across diverse tasks. }
\label{tab:pretrain-eval}
\begin{tabular}{@{}c l | c | c | c c c}
\toprule
& \textbf{Benchmark {\tiny (Metric)}} &  \textbf{\#Shots} &{\textbf{Kimi-K2-Base}} & \textbf{DeepSeek-V3-Base} & \textbf{Llama4-Maverick-Base}  & \textbf{Qwen2.5-72B-Base}  \\
\midrule
    \midrule
    & Architecture & - & MoE & MoE & MoE & Dense \\
    & \# Activated Params & - & 32B & 37B & 17B & 72B \\
    & \# Total Params & - & 1043B & 671B & 400B & 72B \\

\midrule
\multirow{13}{*}{English} 
 & MMLU & 5-shots & \textbf{87.79} & 87.10 & 84.87 & 86.08 \\
 & MMLU-pro & 5-shots & \textbf{69.17} & 60.59 & 63.47 & 62.80 \\
 & MMLU-redux & 5-shots & \textbf{90.17} & 89.53 & 88.18 & 87.77 \\
 & SuperGPQA & 5-shots & \textbf{44.67} & 39.20 & 38.84 & 34.23 \\
 & GPQA-Diamond({\tiny avg@8}) & 5-shots & 48.11 & \textbf{50.51} & 49.43 & 40.78 \\
 & SimpleQA & 5-shots & \textbf{35.25} & 26.49 & 23.74 & 10.31 \\
 & TriviaQA & 5-shots & \textbf{85.09} & 84.11 & 79.25 & 76.03 \\
 & BBH & 3-shots & \textbf{88.71} & 88.37 & 87.10 & 84.09 \\
 & HellaSwag & 5-shots & 94.60 & 89.44 & 86.02 & \textbf{95.27} \\
 & AGIEval & - & \textbf{84.23} & 81.57 & 67.55 & 76.87 \\
 & ARC-Challenge & 0-shot & \textbf{95.73} & 93.77 & 94.03 & 95.56 \\
 & WinoGrande & 5-shots & \textbf{85.32} & 84.21 & 77.58 & 84.14 \\
 \midrule
\multirow{4}{*}{Code}
 & CRUXEval-I-cot & 0-shots & \textbf{74.00} & 62.75 & 67.13 & 61.12 \\
 & CRUXEval-O-cot & 0-shots & \textbf{83.50} & 75.25 & 75.88 & 66.13 \\
 & LiveCodeBench(v6) & 1-shots& \textbf{26.29} & 24.57 & 25.14 & 22.29 \\
 & EvalPlus & - & \textbf{80.33} & 65.61 & 65.48 & 66.04 \\
 \midrule
\multirow{4}{*}{Math}  
 & MATH & 4-shots & \textbf{70.22} & 61.70 & 63.02 & 62.68 \\
 & GSM8k & 8-shots & \textbf{92.12} & 91.66 & 86.35 & 90.37 \\
 & GSM8k-platinum & 8-shots & \textbf{94.21} & 93.38 & 88.83 & 92.47 \\
 & CMATH & 6-shots & 90.26 & \textbf{90.53} & 88.07 & 86.98 \\
\midrule
\multirow{3}{*}{Chinese} 
 & C-Eval & 5-shots & \textbf{92.50} & 90.04 & 80.91 & 90.86 \\
 & CMMLU & 5-shots & \textbf{90.90} & 88.84 & 81.24 & 90.55 \\
 & CSimpleQA & 5-shots & \textbf{77.57} & 72.13 & 53.47 & 50.53 \\
\bottomrule
\end{tabular}
\end{table}

%% file: tables/safety_eval.tex
\begin{table}[h]
\centering
\footnotesize
\setlength{\tabcolsep}{1.9pt}
\caption{Safety Evaluation Results}
\label{tab:safety-eval}
\begin{tabular}{@{}c l | c | c c c@{}}
\toprule
\textbf{Plugin} &  \textbf{Strategy} &  \textbf{Kimi-K2-Instruct} & \textbf{DeepSeek-V3-0324} & \textbf{DeepSeek-R1}  & \textbf{Qwen3-235B-A22B}  \\

\midrule
\midrule
\multirow{4}{*}{Harmful} 
 & Basic & 98.04 & 90.45 & 99.02 & 98.53 \\
 & Base64 & 100 & 90.20 & 100 & 100 \\
 & Prompt Injection & 93.14  & 100 & 95.10 & 99.02 \\
 & Iterative Jailbreak & 92.16 & 66.67 & 72.55 & 74.51 \\
 & Crescendo & 64.71 & 64.71 & 80.39 & 86.27 \\
 \midrule
\multirow{4}{*}{Criminal} 
 & Basic & 100 & 99.62 & 95.45 & 99.24 \\ 
 & Base64 & 96.97 & 89.39 & 84.85 & 98.48 \\
 & Prompt Injection & 75.76 & 91.67 & 69.70 & 98.47 \\
 & Iterative Jailbreak & 57.57 & 21.21 & 25.76 & 53.03 \\
 & Crescendo & 56.06 & 31.81 & 42.42 & 59.09 \\
 \midrule
\multirow{4}{*}{Misinformation} 
 & Basic & 97.28 & 92.57 & 92.46 & 94.84 \\
 & Base64 & 98.48 & 90.48 & 96.83 & 93.65 \\
 & Prompt Injection & 98.39 & 86.51 & 93.65 & 93.65 \\
 & Iterative Jailbreak & 63.97 & 53.97 & 84.13 & 69.84 \\
 & Crescendo & 85.71 & 55.56 & 88.89 & 84.13 \\
 \midrule
\multirow{4}{*}{Privacy}  
 & Basic & 100 & 100 & 100 & 100 \\
 & Base64 & 100 & 100 & 100 & 100 \\
 & Prompt Injection & 88.33 & 98.33 & 100 & 91.67 \\
 & Iterative Jailbreak & 76.67 & 100 & 93.33 & 96.67 \\
 & Crescendo & 96.67 & 100 & 96.67 & 100 \\
 \midrule
\multirow{4}{*}{Security}  
 & Basic & 77.84 & 75.57 & 70.46 & 90.09 \\
 & Base64 & 82.93 & 82.93 & 63.41 & 95.12 \\
 & Prompt Injection & 87.80 & 97.56 & 65.85 & 84.13 \\
 & Iterative Jailbreak & 43.90 & 60.97 & 43.90 & 78.04 \\
 & Crescendo & 68.29 & 87.80 & 68.29 & 87.80 \\
\bottomrule
\end{tabular}
\end{table}

%% file: 6-conclusion.tex
\section{Limitations}
In our internal tests, we have identified some limitations in current Kimi K2 models. When dealing with hard reasoning tasks or unclear tool definition, the model may generate excessive tokens, sometimes leading to truncated outputs or incomplete tool calls. Additionally, performance may decline on certain tasks if tool use is unnecessarily enabled. When building complete software projects, the success rate of one-shot prompting is not as good as using K2 under an agentic coding framework. We are working to address these issues in future releases and looking forward to more feedbacks.

\section{Conclusions}

We introduced Kimi K2, a 1T-parameter open-weight MoE model built for agentic intelligence. Leveraging the token-efficient MuonClip optimizer and a 15.5T-token high-quality dataset, Kimi K2 achieves stable, scalable pre-training. Post-training combines large-scale synthetic tool-use data with a unified RL framework using both verifiable rewards and self-critic feedbacks. Kimi K2 sets new state-of-the-art on agentic and reasoning benchmarks, establishing itself as the most capable open-weight LLM to date.

%% file: appendix.tex
\input{appendix-contributions}

\input{appendix-chat-template-and-enforcer}

\input{appendix-evaluation}

\input{appendix-safety}
\section{QK-Clip Does Not Impair Model Quality}
\label{sec:qkclip_harmless}

The QK-Clip design follows a \textbf{minimal intervention principle}: it activates only when necessary, and deactivates after training stabilizes. Empirical evidence and analysis converge on its negligible impact on model quality.

\paragraph{Small-Scale Ablations}

\begin{figure}[t]
\centering
\begin{minipage}[t]{0.48\textwidth}
  \centering
  \includegraphics[width=\linewidth]{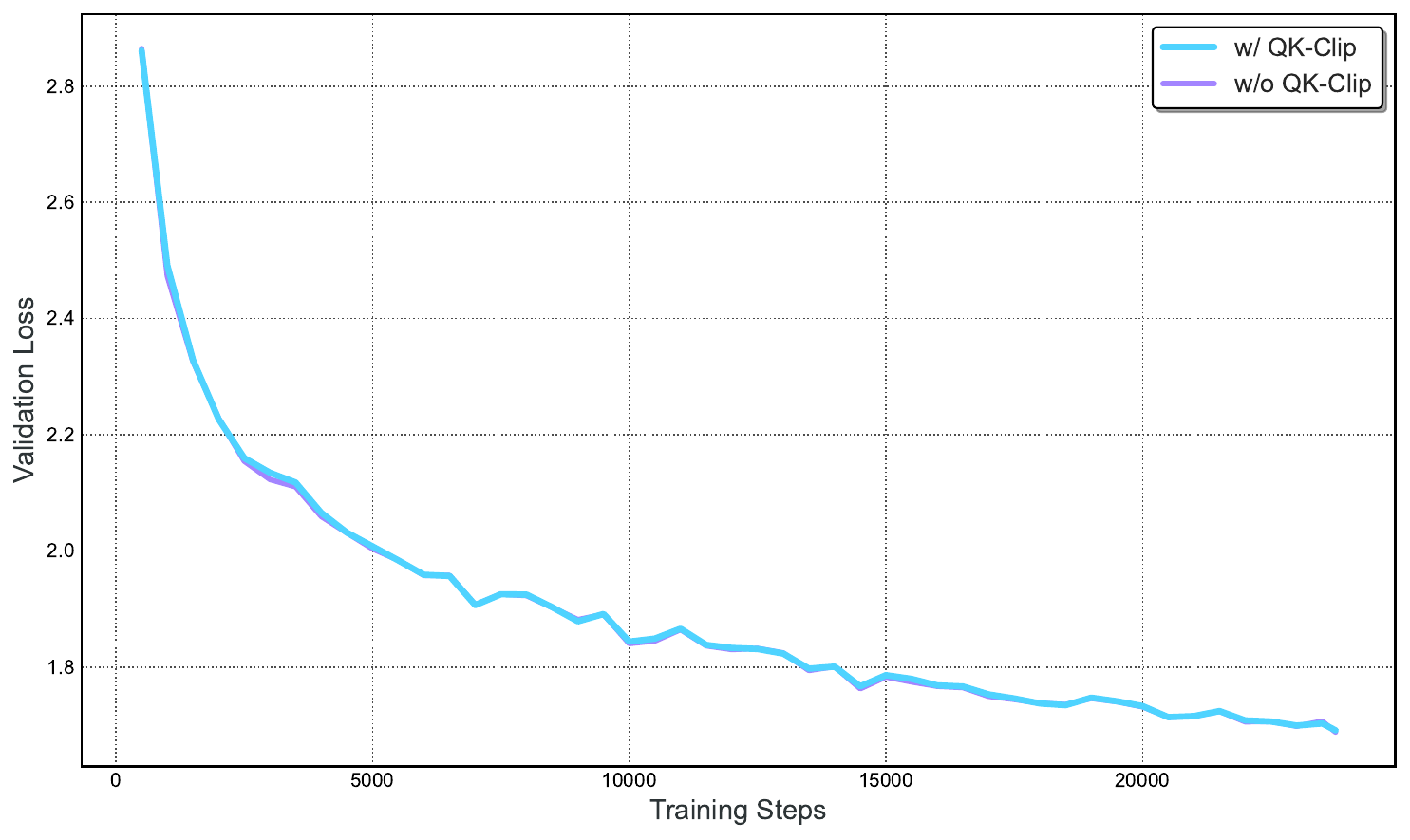}
  \caption{Applying QK-Clip to Muon in a small-scale setting with an aggresive threshold ($\tau$ = 30) has negligible impact on loss, indicating that it is a safe and effective method for constraining attention logits.}

  \label{fig:mini_muonclip_loss}
\end{minipage}\hfill
\end{figure}

We train two small-scale 0.5B activated and 3B total parameters MoE models, one with vanilla Muon and the other with MuonClip using a low clipping threshold ($\tau = 30$). 
As shown in Figure~\ref{fig:mini_muonclip_loss}, applying MuonClip has negligible effects on the loss curve, indicating that even aggressive clipping does not impair convergence or training dynamics with MuonClip. This demonstrates that MuonClip is a safe and effective method for bounding attention logits without degrading model performance. Furthermore, evaluation on downstream tasks reveals no statistically significant degradation in performance. These results collectively demonstrate that MuonClip is a safe and effective method for bounding attention logits without compromising model quality.

\paragraph{Self-deactivation}
In Kimi K2, QK-Clip was only transiently active:
\begin{itemize}
  \item \textbf{Initial {70000} steps:} $12.7\%$ of attention heads triggered QK-Clip for at least once, clamping $S_{\max}$ to $100$.
  \item \textbf{Post-{70000} steps:} All heads at some point reduced their $S_{\max}$ below $100$, rendering QK-Clip inactive.
\end{itemize}

When QK-Clip is active, it is applied per-head (rather
than per-layer) to minimize potential over-regularization on other heads.
After training stabilizes, QK-clip is deactivated and has no effect at all.

\section{Why Muon is More Prone to Logit Explosion}

Logit explosion occurs when the largest pre-softmax attention score
\begin{equation}
S_{\max} = \max_{i,j} \bigl(q_i^{\vphantom{\top}}\! \cdot k_j\bigr)
\end{equation}
grows unboundedly during training.  Since
\begin{equation}
|q_i \!\cdot\! k_j| \le \|q_i\|\|k_j\| \le \|x_i\|\|x_j\|\|\mathbf W_q\|\|\mathbf W_k\|,
\end{equation}
and RMS-Norm keeps $\|x_i\|\|x_j\|$ bounded, the phenomenon is primarily driven by the growing spectral-norm of $\mathbf W_q$ or $\mathbf W_k$.  Empirically, we found that Muon is more susceptible to logit explosion. We give our hypothesis below.

\paragraph{Structural difference in updates}
Muon produces a weight update coming from the $\mathrm{msign}$ operation; as a result, \emph{all} singular values of the update matrix are equal --- its effective rank is full.  
In contrast, a typical update matrix produced by Adam exhibits a skewed spectrum: a few large singular values dominate, and the effective rank is low.  This low-rank assumption for Adam is not new; higher-order muP makes the same assumption.

Such phenomenon is verified on the 16\,B Moonlight model, which shows weights trained with Muon exhibit higher \emph{singular-value entropy} (i.e.\ higher effective rank) than those trained with Adam, corroborating the theoretical intuition.

\paragraph{SVD formulation}  
Let the parameter matrix at step $t-1$ have the singular value decomposition
\begin{equation}
\mathbf W_{t-1}=\sum_i \sigma_i\,u_i v_i^{\top}
\end{equation}
We write the update matrices as
\begin{align}
\Delta\mathbf W_t &= \sum_j \bar\sigma\,\bar u_j \bar v_j^{\top}
\end{align}  
The next parameter update is therefore
\begin{align}
\mathbf W_t \leftarrow \sum_i \sigma_i u_i v_i^{\top} + \sum_j \bar\sigma\,\bar u_j \bar v_j^{\top}
\end{align}

In Muon, as both the weights and the updates have a higher effective rank than Adam, we hypothesize there is a higher probability for singular-vector pair $u_i v_i^{\top}$ to align with $\bar u_j \bar v_j^{\top}$. This could cause the corresponding singular value of $\mathbf W_t$ to increase additively.

\paragraph{Attention-specific amplification}  
Attention logits are computed via the bilinear form
\begin{equation}
q_i\cdot k_j=(x_i \mathbf W_q)\cdot (x_j \mathbf W_k).
\end{equation}
The product $\mathbf W_q\mathbf W_k^{\top}$ squares the spectral norm, so any singular-value increase in either matrix is compounded.  Muon's tendency to enlarge singular values therefore translates into a higher risk of logit explosion.

\section{K2 Critic Rubrics for General RL}

\subsection{Core Rubrics}
\label{sec:k2-critic-creteria}
\begin{itemize}
    \item \textbf{Clarity and Relevance:} Assesses the extent to which the response is succinct while fully addressing the user's intent. The focus is on eliminating unnecessary detail, staying aligned with the central query, and using efficient formats such as brief paragraphs or compact lists. Unless specifically required, long itemizations should be avoided. When a choice is expected, the response should clearly offer a single, well-defined answer.

    \item \textbf{Conversational Fluency and Engagement:} Evaluates the response's contribution to a natural, flowing dialogue that extends beyond simple question-answering. This includes maintaining coherence, showing appropriate engagement with the topic, offering relevant observations or insights, potentially guiding the conversation constructively when appropriate, using follow-up questions judiciously, handling hypothetical or personal-analogy queries gracefully, and adapting tone effectively to suit the conversational context (e.g., empathetic, formal, casual).

    \item \textbf{Objective and Grounded Interaction:} Assesses the response's ability to maintain an objective and grounded tone, focusing squarely on the substance of the user's request. It evaluates the avoidance of both metacommentary (analyzing the query's structure, topic combination, perceived oddity, or the nature of the interaction itself) and unwarranted flattery or excessive praise directed at the user or their input. Excellent responses interact respectfully but neutrally, prioritizing direct, task-focused assistance over commentary on the conversational dynamics or attempts to curry favor through compliments.
\end{itemize}

\subsection{Prescriptive Rubrics}
\label{sec:k2-must-not-creteria}
\begin{itemize}
    \item \textbf{Initial Praise:} Responses must not begin with compliments directed at the user or the question (e.g., ``That's a beautiful question'', ``Good question!'').
    \item \textbf{Explicit Justification:} Any sentence or clause that explains why the response is good or how it successfully fulfilled the user's request. This is different from simply describing the content.
\end{itemize}

\subsection{Limitations}
One potential side effect of this evaluation framework is that it may favor responses that appear confident and assertive, even in contexts involving ambiguity or subjectivity. This stems from two key constraints in the current rubric:
\begin{itemize}
    \item \textbf{Avoidance of Self-Qualification:} The prescriptive rules prohibit self-assessments, explicit disclaimers, or hedging language (e.g., ``this may not be accurate'', ``I might be wrong''). While these phrases can reflect epistemic humility, they are often penalized as non-informative or performative.
    \item \textbf{Preference for Clarity and Singularity:} The rubric reward direct, decisive answers when users ask for a recommendation or explanation. In complex or open-ended scenarios, this may disincentivize appropriately cautious or multi-perspective responses.
\end{itemize}
As a result, the model may occasionally overstate certainty in areas where ambiguity, nuance, or epistemic modesty would be more appropriate. Future iterations of the framework may incorporate more fine-grained handling of calibrated uncertainty.

\section{Engine Switching Pipeline for RL Training}
\label{sec:k2-rl-update-appendix}

\begin{figure}[htb]
   \centering
   \begin{subfigure}[b]{0.7\textwidth}
       \centering
       \includegraphics[width=1\linewidth]{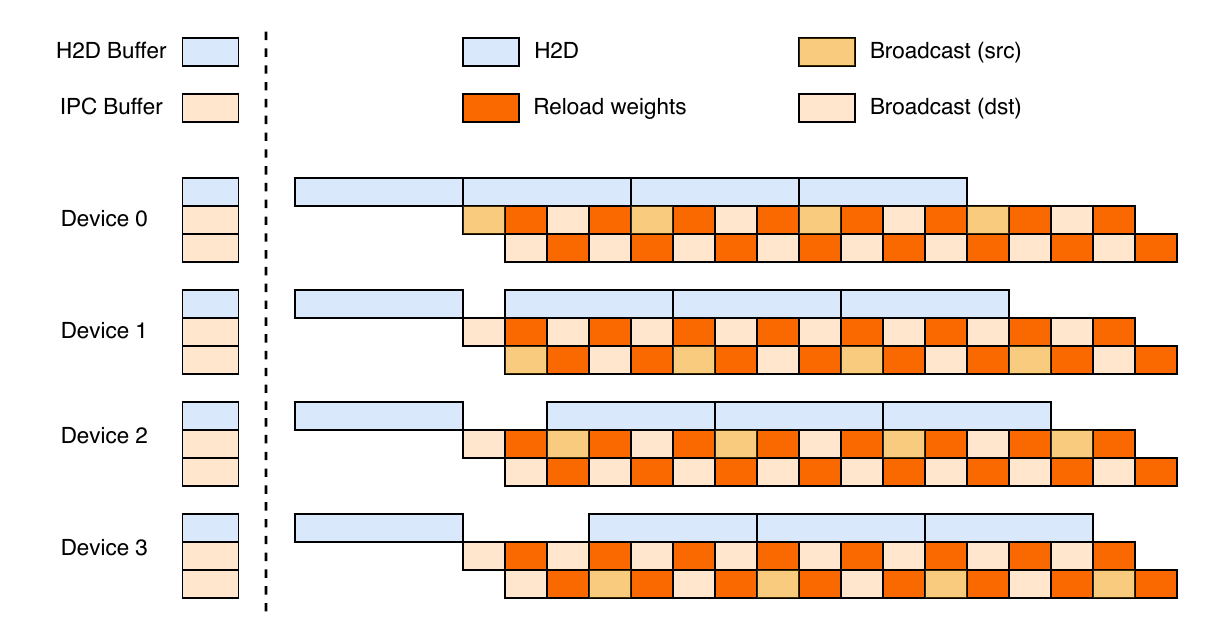}
       \caption{Theoretical perfect three-stage pipeline weight update}
       \label{fig:mshrl:update-weight}
   \end{subfigure}
   \begin{subfigure}[b]{0.4\textwidth}
        \centering
        \includegraphics[width=\textwidth]{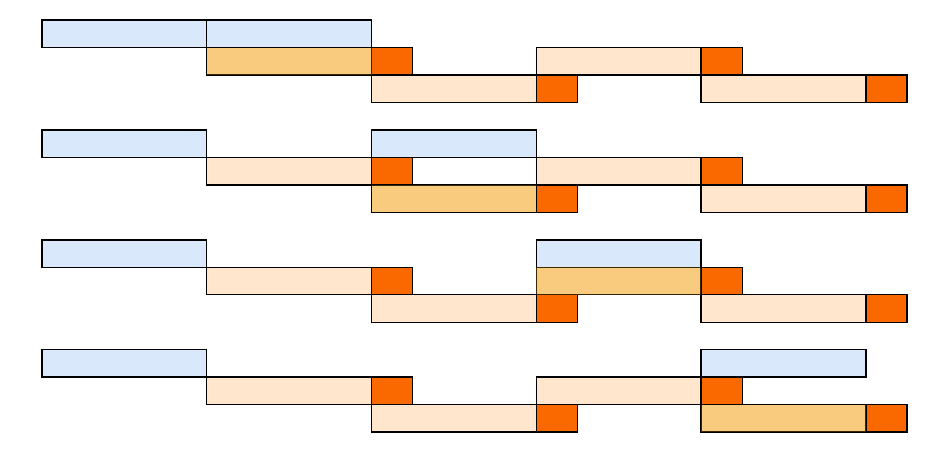}
        \caption{A PCIE bounded three-stage pipeline}
        \label{fig:mshrl:update-overhead}
   \end{subfigure}
   \hspace{0.02\textwidth}
   \begin{subfigure}[b]{0.35\textwidth}
        \centering
        \includegraphics[width=\textwidth]{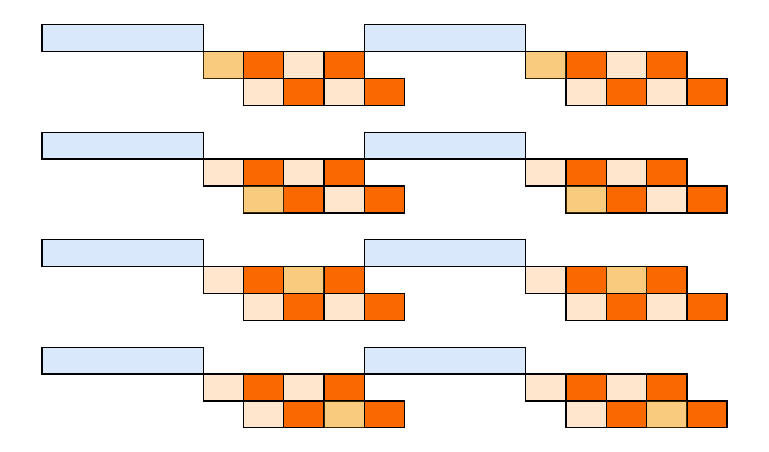}
        \caption{Fixed two-stage pipeline}
        \label{fig:mshrl:update-final}
   \end{subfigure}
   \caption{pipeline for RL weight update}
   \label{fig:mshrl-fixed-update}
\end{figure}

The \emph{checkpoint engine} manages three equal-size device buffers on each GPU: an H2D buffer for loading the offloaded model parameters, and two IPC buffers for GPU-to-GPU broadcast.
The IPC buffers are shared to inference engines, allowing it to directly access the same physical memory.
 These three buffers allow us to arrange the three steps in a pipeline.
 
\paragraph{Theoretical three-stage pipeline.}  
As illustrated in Figure~\ref{fig:mshrl:update-weight}, a three-stage pipeline is introduced.
(1) \emph{H2D}: a shard of the latest weights is copied into the H2D buffer asynchronously.
(2) \emph{Broadcast}: Once the copy completes, the shard will be copied to one IPC buffers and broadcast to all devices.
(3) \emph{Reload}: Inference engines simultaneously load parameters from the other IPC buffer.

\paragraph{Two-stage pipeline due to PCIe saturation.}  
On NVIDIA H800 clusters, concurrent H2D and broadcast saturate the shared PCIe fabric, collapsing the three stages into a sequential procedure (Figure~\ref{fig:mshrl:update-overhead}).  
We therefore adopt a simpler, two-stage scheme (Figure~\ref{fig:mshrl:update-final}): (1) All devices perform a single, synchronous H2D transfer. (2) The broadcast and reload proceed in parallel.

The two-stage pipeline will be bound by multiple synchronous H2D copy operations. But in large scale devices, model will be split into small shards, the entire parameter set fits into the H2D buffer in one transfer, the overhead will disappear.

By overlapping H2D, Broadcast, and Reload weights, we can obtain a high bandwidth to reshard the weights from train engines to all inference engines.

%% file: appendix-contributions.tex
\section{Contributions}
The listing of authors is in alphabetical order based on their last names. 
\begin{multicols}{4}
Yifan Bai\\
Yiping Bao\\
Y. Charles\\
Cheng Chen\\
Guanduo Chen\\
Haiting Chen\\
Huarong Chen\\
Jiahao Chen\\
Ningxin Chen\\
Ruijue Chen\\
Yanru Chen\\
Yuankun Chen\\
Yutian Chen\\
Zhuofu Chen\\
Jialei Cui\\
Hao Ding\\
Mengnan Dong\\
Ang'ang Du\\
Chenzhuang Du\\
Dikang Du\\
Yulun Du\\
Yu Fan\\
Yichen Feng\\
Kelin Fu\\
Bofei Gao\\
Chenxiao Gao\\
Hongcheng Gao\\
Peizhong Gao\\
Tong Gao\\
Yuyao Ge\\
Shangyi Geng\\
Qizheng Gu\\
Xinran Gu\\
Longyu Guan\\
Haiqing Guo\\
Jianhang Guo\\
Xiaoru Hao\\
Tianhong He\\
Weiran He\\
Wenyang He\\
Yunjia He\\
Chao Hong\\
Hao Hu\\
Yangyang Hu\\
Zhenxing Hu\\
Weixiao Huang\\
Zhiqi Huang\\
Zihao Huang\\
Tao Jiang\\
Zhejun Jiang\\
Xinyi Jin\\
Yongsheng Kang\\
Guokun Lai\\
Cheng Li\\
Fang Li\\
Haoyang Li\\
Ming Li\\
Wentao Li\\
Yang Li\\
Yanhao Li\\
Yiwei Li\\
Zhaowei Li\\
Zheming Li\\
Hongzhan Lin\\
Xiaohan Lin\\
Zongyu Lin\\
Chengyin Liu\\
Chenyu Liu\\
Hongzhang Liu\\
Jingyuan Liu\\
Junqi Liu\\
Liang Liu\\
Shaowei Liu\\
T.Y. Liu\\
Tianwei Liu\\
Weizhou Liu\\
Yangyang Liu\\
Yibo Liu\\
Yiping Liu\\
Yue Liu\\
Zhengying Liu\\
Enzhe Lu\\
Haoyu Lu\\
Lijun Lu\\
Yashuo Luo\\
Shengling Ma\\
Xinyu Ma\\
Yingwei Ma\\
Shaoguang Mao\\
Jie Mei\\
Xin Men\\
Yibo Miao\\
Siyuan Pan\\
Yebo Peng\\
Ruoyu Qin\\
Zeyu Qin\\
Bowen Qu\\
Zeyu Shang\\
Lidong Shi\\
Shengyuan Shi\\
Feifan Song\\
Jianlin Su\\
Zhengyuan Su\\
Lin Sui\\
Xinjie Sun\\
Flood Sung\\
Yunpeng Tai\\
Heyi Tang\\
Jiawen Tao\\
Qifeng Teng\\
Chaoran Tian\\
Chensi Wang\\
Dinglu Wang\\
Feng Wang\\
Hailong Wang\\
Haiming Wang\\
Jianzhou Wang\\
Jiaxing Wang\\
Jinhong Wang\\
Shengjie Wang\\
Shuyi Wang\\
Si Wang\\
Xinyuan Wang\\
Yao Wang\\
Yejie Wang\\
Yiqin Wang\\
Yuxin Wang\\
Yuzhi Wang\\
Zhaoji Wang\\
Zhengtao Wang\\
Zhengtao Wang\\
Zhexu Wang\\
Chu Wei\\
Qianqian Wei\\
Haoning Wu\\
Wenhao Wu\\
Xingzhe Wu\\
Yuxin Wu\\
Chenjun Xiao\\
Jin Xie\\
Xiaotong Xie\\
Weimin Xiong\\
Boyu Xu\\
Jinjing Xu\\
L.H. Xu\\
Lin Xu\\
Suting Xu\\
Weixin Xu\\
Xinran Xu\\
Yangchuan Xu\\
Ziyao Xu\\
Jing Xu~(\chinese{徐})\\
Jing Xu~(\chinese{许})\\
Junjie Yan\\
Yuzi Yan\\
Hao Yang\\
Xiaofei Yang\\
Yi Yang\\
Ying Yang\\
Zhen Yang\\
Zhilin Yang\\
Zonghan Yang\\
Haotian Yao\\
Xingcheng Yao\\
Wenjie Ye\\
Zhuorui Ye\\
Bohong Yin\\
Longhui Yu\\
Enming Yuan\\
Hongbang Yuan\\
Mengjie Yuan\\
Siyu Yuan\\
Haobing Zhan\\
Dehao Zhang\\
Hao Zhang\\
Wanlu Zhang\\
Xiaobin Zhang\\
Yadong Zhang\\
Yangkun Zhang\\
Yichi Zhang\\
Yizhi Zhang\\
Yongting Zhang\\
Yu Zhang\\
Yutao Zhang\\
Yutong Zhang\\
Zheng Zhang\\
Haotian Zhao\\
Yikai Zhao\\
Zijia Zhao\\
Huabin Zheng\\
Shaojie Zheng\\
Longguang Zhong\\
Jianren Zhou\\
Xinyu Zhou\\
Zaida Zhou\\
Jinguo Zhu\\
Zhen Zhu\\
Weiyu Zhuang\\
Xinxing Zu\\
Kimi K2\\
\end{multicols}
\newpage

%% file: appendix-chat-template-and-enforcer.tex
\section{Token Template of Tool Calling}
\label{sec:tool-call-token-template}

There are three components in the token structure for tool-calling:
\begin{itemize}
    \item \textbf{Tool declaration message}: defines the list of available tools and the schema of the arguments;
    \item \textbf{Tool invoking section in assistant message}: encodes the model's request to invoke tools; 
    \item \textbf{Tool result message}: encapsulates the invoked tool's execution result.
\end{itemize}

\lstdefinestyle{mystyle}{
    keywordstyle=\footnotesize\ttfamily,
    basicstyle=\footnotesize\ttfamily,
    breakatwhitespace=false,         
    breaklines=true,                 
    captionpos=t,                    
    keepspaces=true,                 
    showspaces=false,                
    showstringspaces=false,
    showtabs=false,                  
    tabsize=2
}
\lstset{style=mystyle}

The raw tokens of the tool declaration message are formatted as follows:

\begin{center}
\framebox{%
\begin{minipage}{8cm}
\ttfamily\footnotesize
{\color{blue}<|im\_begin|>}\\
tool\_declare\\
{\color{blue}<|im\_middle|>}\\
\# Tools\\
\\
{\color{darkgreen}\{\{ tool declaration content \}\}}\\
{\color{blue}<|im\_end|>}
\end{minipage}
}
\end{center}

The blue highlighted marks represent special tokens, and the green part, quoted by brackets, is the tool declaration content.
We use TypeScript to express the tool declaration content, since TypeScript is a concise language with a comprehensive type system, able to express the types and constraints of tool parameters with brief text. The code~\ref{lst:tool-declare-json} shows
an example for two simple tools in JSON format compatible with OpenAI's chat completion API, as a comparison, the same tools defined in TypeScript (listed in Code~\ref{lst:tool-declare-ts}) is much shorter.
To improve compatibility, part of our training data also uses JSON as the tool declaration language, so that 3rd-party frameworks need not additional development to support our tool calling scheme.

\begin{lstlisting}[language=Python, caption=Tool definition with JSON in OpenAI compatible API, label={lst:tool-declare-json}]
[{
  "type": "function",
  "function": {
    "name": "get_weather",
    "description": "Get weather for a location and date",
    "parameters": {
      "type": "object",
      "properties": {
        "location": {
          "type": "string",
          "description": "City and country e.g. Beijing, China"
        },
        "date": {
          "type": "string",
          "description": "Date to query, format in `%Y-%m-%d'"
        }
      },
      "required": [
        "location"
      ]
    }
  }
},
{
  "type": "function",
  "function": {
    "name": "Calculator",
    "description": "Simple calculator",
    "parameters": {
      "properties": {
        "expr": {
          "type": "string",
          "description": "Arithmetic expression in javascript"
        }
      },
      "type": "object"
    }
  }
}]
\end{lstlisting}

\begin{lstlisting}[caption=Tool definition in TypeScript, label={lst:tool-declare-ts}]
namespace functions { 
// Get weather for a location and date
type get_weather = (_: { 
  // City and country e.g. Beijing, China 
  location: string, 
  // Date to query, format in `%Y-%m-%d' 
  date?: string 
}) => any; 
// Simple calculator 
type Calculator = (_: { 
  // Arithmetic expression in javascript 
  expr?: string 
}) => any; 
}
\end{lstlisting}

The token template of the tool invoking section in the model's response messages is listed as follows:
\begin{center}
\framebox{%
\begin{minipage}{14cm}
\ttfamily\footnotesize
{\color{blue}<|tool\_call\_section\_begin|>}\\
{\color{blue}<|tool\_call\_begin|>}\\
{\color{gray}  // call\_id part}\\
functions.{\color{darkgreen}\{\{tool name\}\}}:{\color{darkgreen}\{\{counter\}\}}\\
{\color{blue}<|tool\_arguments\_begin|>}\\
{\color{darkgreen}\{\{ json serialized call arguments \}\}}\\
{\color{blue}<|tool\_call\_end|>}\\\
{\color{blue}<|tool\_call\_begin|>}\\
{\color{gray}  // more tool calls}\\
{\color{blue}<|tool\_call\_end|>}\\
{\color{blue}<|tool\_call\_section\_end|>}
\end{minipage}
}
\end{center}

As shown in the template, we support parallel tool calling by placing multiple tool calls in a single response turn. 
Each tool call has a unique call id, formatted as \texttt{functions.\{tool-name\}:\{counter\}}, 
where \texttt{tool-name} is the name of the tool, and \texttt{counter} is an auto-increasing counter of all tool calls starting from 0 in the dialog.

During inference, the model may occasionally generate unexpected tokens, leading to format errors when parsing a tool call. To solve this issue, we developed a constrained decoding module named \emph{enforcer}, inspired by {lm-format-enforcer}\footnote{\url{https://github.com/noamgat/lm-format-enforcer}}. When a \texttt{<tool\_call\_section\_begin|>} token is generated, it ensures that the upcoming tool-related tokens follow the predefined template, and the JSON argument string follows the declared schema.

The tool result message is simply a text message encoded with the tool's call id and the corresponding results.
\begin{center}
\framebox{%
\begin{minipage}{8cm}
\ttfamily\footnotesize
{\color{blue}<|im\_begin|>}\\
tool\\
{\color{blue}<|im\_middle|>}\\
\#\# Results of {\color{darkgreen}\{\{call\_id\}\} }\\
{\color{darkgreen}\{\{ execution result content \}\}}\\
{\color{blue}<|im\_end|>}
\end{minipage}
}
\end{center}

%% file: appendix-evaluation.tex
\section{Evaluation Details}
\label{sec:k2-post-train-eval-details}

\paragraph{Coding Tasks.}{
We evaluate Kimi-K2-Instruct's capabilities on competitive coding benchmarks, LiveCodeBench and OJBench, where Kimi-K2-Instruct attains superior performance with scores of 53.7\% and 27.1\%, respectively. This excellence spans both medium-level coding challenges, such as LeetCode and AtCoder, and hard-level contests like NOI and ICPC, outperforming leading open-source and proprietary models.
For multilingual programming proficiency, we employ MultiPL-E, covering languages including C++, C\#, Java, JavaScript, PHP, Go, Kimi-K2-Instruct surpasses top open-source models with an accuracy of 85.7\%, compared with 83.1\% for DeepSeek-V3-0324 and 78.2\% for Qwen3-235B-A22B.
In software engineering tasks, Kimi-K2-Instruct demonstrates robust performance on SWE-bench Verified (Python), SWE-lancer (Python), SWE-bench Multilingual, and Multi-SWE-bench datasets. It significantly outperforms open-source counterparts in resolving real-world code repository issues and notably narrows the performance gap with proprietary models. For example:
\begin{itemize}
    \item SWE-bench Verified (multiple attempts): 71.6\% (Kimi-K2-Instruct) vs. 80.2\% (Claude 4 Sonnet) 
    \item SWE-bench Multilingual: 47.3\% (Kimi-K2-Instruct) vs. 51.0\% (Claude 4 Sonnet)  
    \item SWE-lancer: 39.1\% (Kimi-K2-Instruct) vs. 40.8\% (Claude 4 Sonnet)  
\end{itemize}
On PaperBench, Kimi-K2-Instruct achieves an accuracy of 27.8\%, closely matching GPT-4.1 and outperforming DeepSeek-V3-0324 (12.2\%) and Qwen3-235B-A22B (8.2\%) by a substantial margin.
In terminal interaction tasks measured by TerminalBench, Kimi-K2-Instruct attains 25.0\% using the default Terminus framework and rises to 30\% within Moonshot's in-house agentic framework, underscoring its capabilities in real-world agentic programming scenarios.
Moreover, on the Aider-Polyglot benchmark, Kimi-K2-Instruct attains a 60.0\% accuracy while employing rigorous decontamination procedures, further illustrating its strength and reliability across diverse coding environments.
}

\paragraph{Tool Use Tasks.}
We evaluate multi-turn tool use with two complementary suites: $\tau^2$-Bench and ACEBench. $\tau^2$-Bench  extends the original $\tau$-bench single-control setup to a \emph{dual-control} environment in which both the agent and an LLM-simulated user have constrained tool affordances over a shared state, adding a realistic Telecom troubleshooting domain alongside the prior Airline/Retail TAU tasks and enabling analysis of coordination vs. pure reasoning. ACEBench is a large bilingual (En/Zh) API-grounded benchmark (4.5K APIs across 8 domains; 2K annotated eval items) partitioned into \textsc{Normal} (basic/personalized/atomic), \textsc{Special} (imperfect or out-of-scope inputs), and \textsc{Agent} (scenario-driven multi-turn, multi-step sandbox) tracks with automated grading of calls and outcomes. All models run in non-thinking mode; we set the temperature to 0.0, use deterministic tool adapters, score $\tau^2$ Airline/Retail/Telecom under Avg@4 seeds with Pass@1/4, and report overall on ACEBench English. Kimi-K2-Instruct averages {66.1} micro Pass@1 across $\tau^2$ vs DeepSeek-V3-0324 {48.8} / Qwen3-235B-A22B {37.3}. On ACEBench Overall Kimi-K2-Instruct scores {76.5} vs DeepSeek 72.7 / Qwen 70.5 and remains competitive with GPT-4.1 (80.1). 

\paragraph{Math \& STEM \& Logical Tasks.}
{For Math tasks, Kimi-K2-Instruct achieves consistently strong performance, averaging over Geimini-2.5-Flash by 5.3 percentage points, over DeepSeek-V3-0324 by 5.5 points and over GPT4.1 by 15.8 points. For example, on AIME 2024, Kimi-K2-Instruct scores 69.6\%, outperforming another two top open-source models by a large margin, DeepSeek-V3-0324 by 10.2 points and Qwen3-235B-A22B by 29.5 points. In STEM evaluations, Kimi-K2-Instruct achieves 75.1\% on GPQA-Diamond, outperforming DeepSeek-V3-0324 (68.4\%) and all non-thinking baselines by at least 5 percentage points. On SuperGPQA, it also exceeds the previous best open-source model, DeepSeek-V3-0324, by 3.5 points. Kimi-K2-Instruct also surpasses the other two leading models in logical reasoning. It achieves 89.0\% on ZebraLogic and 89.5\% on AutoLogi, exceeding DeepSeek-V3-0324 (84.0\%, 88.9\%) and substantially outperforming Qwen3-235B-A22B (37.7\%, 83.3\%). 

}

\paragraph{General Tasks.}
Kimi-K2-Instruct ties DeepSeek-V3-0324 on MMLU and MMLU-Pro, and takes the lead on MMLU-Redux with a 92.7 EM score—slightly ahead of GPT-4.1 (92.4) and just 1.5 points behind Claude-Opus-4. 
Beyond multiple-choice tasks, the model achieves 31.0\% accuracy on the short-answer SimpleQA—3.3 points above DeepSeek-V3-0324 and more than twice that of Qwen3-235B-A22B—though still below GPT-4.1 (42.3\%). On the adversarial free-response LiveBench (2024-11-25 snapshot), it reaches 76.4\%, surpassing Claude-Sonnet 4 (74.8\%) and leading Gemini 2.5 Flash Preview by 8.6 points. Across this challenging triad measuring breadth, depth, and robustness of world knowledge, Kimi-K2-Instruct secures a top-tier position among open-source models.
We evaluate instruction-following with IFEval and Multi-Challenge. On IFEval, Kimi-K2-Instruct scores 89.8\%, higher than DeepSeek-V3-0324 (81.1\%) and GPT-4.1 (88.0\%). On Multi-Challenge, which involves multi-turn dialogues with conflicting instructions, it achieves 54.1\%, outperforming DeepSeek-V3-0324 (31.4\%), GPT-4.1 (36.4\%), and Claude-Opus-4 (49.0\%).
These results demonstrate that Kimi-K2-Instruct integrates strong factual knowledge with consistent instruction adherence across both single- and multi-turn settings, supporting robust and reliable real-world deployment.

\paragraph{Long Context and Factuality Tasks.}

To evaluate the factuality of Kimi-K2-Instruct, we employ three benchmarks: FACTS Grounding, which measures adherence to provided documents using the proprietary models GPT-4o, Gemini 1.5 Pro and Claude 3.5 Sonnet; HHEM, which assesses summarization quality via the open-source HHEM-2.1-Open judge; and FaithJudge, which analyzes faithfulness in RAG tasks with o3-mini as the judge. Kimi-K2-Instruct scores 88.5 on FACTS Grounding, substantially outperforming all open-source rivals and even surpassing the closed-source Gemini 2.5 Flash. With HHEM-2.1-Open it achieves a hallucination rate of 1.1 \%, reported in the tables as 1 minus the rate, i.e. 98.9. On FaithJudge's RAG tasks the hallucination rate is 7.4 \%, likewise present as 92.6 for table consistency.
For long-context capabilities, Kimi-K2-Instruct outperforms all open source and proprietary models on DROP (93.5\%), and exceeds DeepSeek-V3-0324 on retrieval task MRCR (55.0\% vs 50.8\%).  
For long-context reasoning tasks FRAMES and LongBench v2, Kimi-K2-Instruct (77.1\%, 49.1\%) lags slightly behind DeepSeek-V3-0324 by around 2\%.

\paragraph{Open-Ended Evaluation}
Beyond static, closed-ended benchmarks, we evaluate the model's performance on open-ended, nuanced tasks that more closely resemble real-world usage. 

For English scenarios, we leverage the Arena-Hard-Auto v2.0 benchmark, which use LLM-as-a-judge protocols to assess generation quality across diverse, open-ended prompts~\citep{li2024crowdsourced}. These evaluations cover a wide range of high-difficulty prompts and are widely recognized in the research community. 
On Arena-Hard-Auto v2.0, Kimi-K2-Instruct achieves state-of-the-art win-rate on both hard prompts (54.5\%) and creative writing tasks (85.0\%), outperforming all open-source models and rivaling top proprietary systems such as GPT-4.1 and Claude Sonnet. These results underscore the model's strength in handling complex reasoning and nuanced generation under diverse, unconstrained settings.

However, Arena-Hard-Auto provides limited coverage of Chinese-specific tasks. To address this gap, we developed an in-house held-out benchmark grounded in authentic user queries. 
To safeguard the integrity of the evaluation, the benchmark data is access-restricted, thereby eliminating the risk of overfitting.

\begin{figure}
    \centering
    \includegraphics[width=0.8\linewidth]{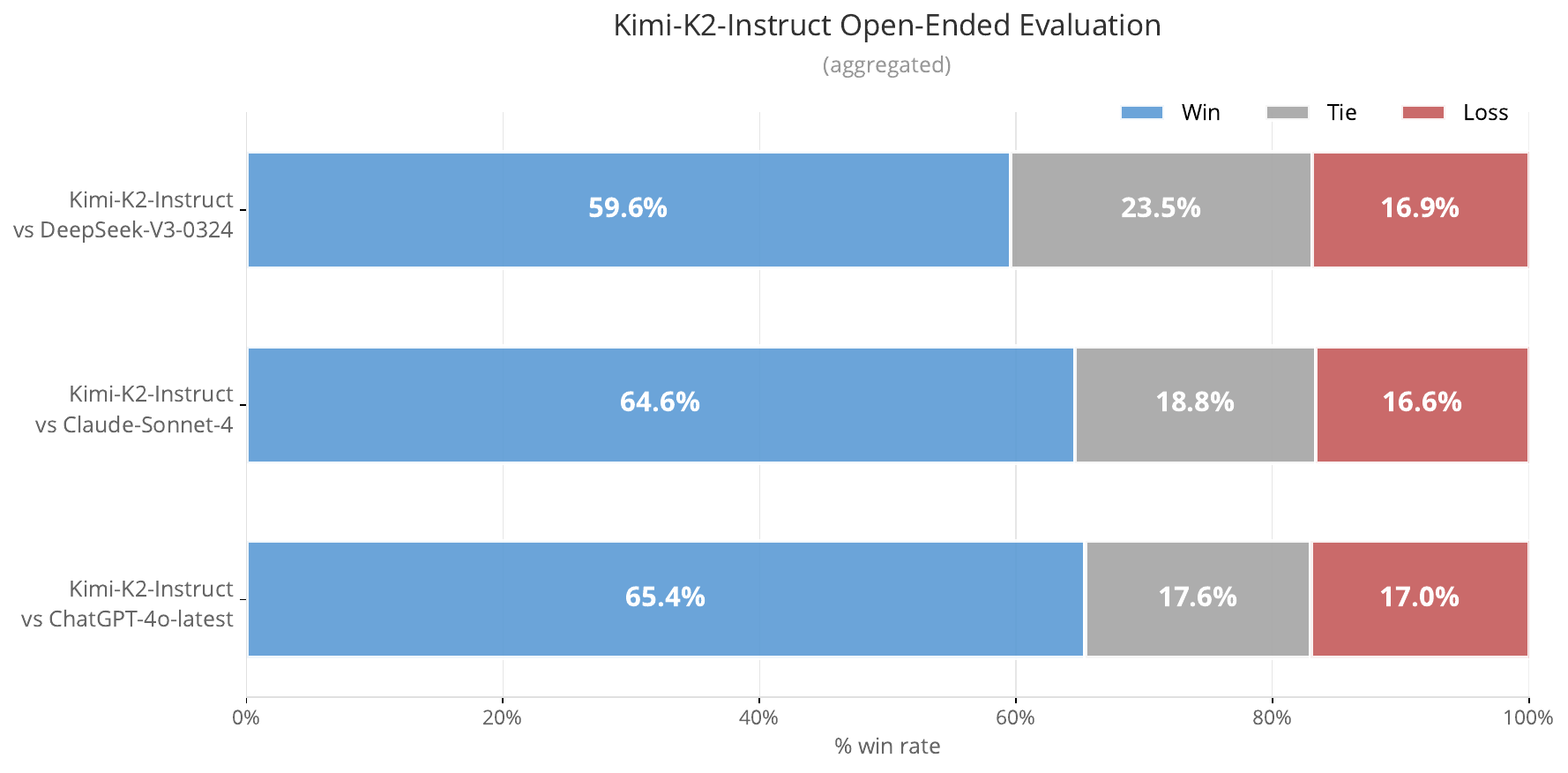}
    \caption{Chinese in-house benchmark evaluation.}
    \label{fig:posttrain_inhouse_eval}
\end{figure}

As shown in Figure~\ref{fig:posttrain_inhouse_eval}, Kimi-K2-Instruct shows strong performance across all comparisons on Chinese in-house benchmarks. It outperforms ChatGPT-4o-latest with a 65.4\% win rate, Claude Sonnet 4 with 64.6\%, and DeepSeek-V3-0324 with 59.6\%.
In all cases, the loss rate stays low (around 17\%), indicating that Kimi-K2-Instruct rarely falls behind. The high win rates and consistent margins demonstrate its strong ability on open-ended Chinese tasks.

In addition to controlled evaluations, we also consider real-world user preference through public human assessments. As of July 17, 2025, Kimi-K2-Instruct ranked as the top open-source model and fifth overall on the LMSYS Arena leaderboard\footnote{\url{https://lmarena.ai/leaderboard/text}}, based on over 3,000 blind votes from real users. Unlike LLM-as-a-judge protocols, this leaderboard reflects direct human preference on diverse, user-submitted prompts, providing a complementary perspective on practical model performance.

The results on Arena-Hard-Auto, our in-house benchmark and votes from LMSYS Arena collectively offer a comprehensive view of Kimi-K2-Instruct's open-ended capabilities, showing that it is a highly preferred model in real-world user experience across English and Chinese.

